\documentclass[lualatex,16pt,a4paper,table]{article}
\usepackage{booktabs}
\usepackage{colortbl} 
\usepackage[table]{xcolor}
\definecolor{pelicanmain}{HTML}{FBE6B2}
\definecolor{pelicanlight}{HTML}{FFF3D6}
\usepackage{xspace}
\usepackage{tabularx} 
\usepackage{booktabs} 
\usepackage{graphicx}
\usepackage{natbib}
\usepackage{amssymb}
\usepackage{arpafvg}
\usepackage{caption,booktabs}
\usepackage{lipsum}
\usepackage{subcaption}
\usepackage[most]{tcolorbox}
\definecolor{mylightgreen}{HTML}{D7FFD7}
\definecolor{myred}{RGB}{255,0,0}
\definecolor{modiblue}{RGB}{0,40,161}
\usepackage{lipsum}      
\usepackage{booktabs} 
\usepackage{multirow} 
\usepackage{caption} 
\usepackage{makecell}
\usepackage{adjustbox} 
\newtcolorbox{takeawaybox}{
  colback=gray!10,  
  colframe=gray!75, 
  boxrule=0.5pt,    
  arc=2mm,          
  halign=center,    
  valign=center,    
  boxsep=3pt,       
  fontupper=\small\bfseries\sffamily, 
  sharp corners,    
}
\newtcolorbox{casebox}[1][]{
    enhanced,
    colback=black!5!white, 
    colframe=black!60!white, 
    boxrule=0.8pt,           
    arc=2mm,                 
    fonttitle=\bfseries\sffamily, 
    title=#1,                
    left=6pt, right=6pt, top=6pt, bottom=6pt, 
    sharp corners,
    breakable,               
    #1 
}

\newcommand{\model}{Pelican-VL 1.0\xspace}

\usepackage{float} 

\begin{document}

\makeatletter

\title{\textcolor{titlecolor}{Pelican-VL 1.0: A Foundation Brain Model for Embodied Intelligence}}

\vspace{10em}

\author{\textbf{WFM System Group} \\ 
	Beijing Innovation Center of Humanoid Robotics (X-Humanoid) \\
	\textit{Pelican-VL.github.io} \\
    \{vito.dai, jason.ju\}@x-humanoid.com
    }
\date{\today}

\maketitle

\makeatother

\pagebreak[4]




\begin{figure}
    \centering
    \includegraphics[width=1.05\linewidth]{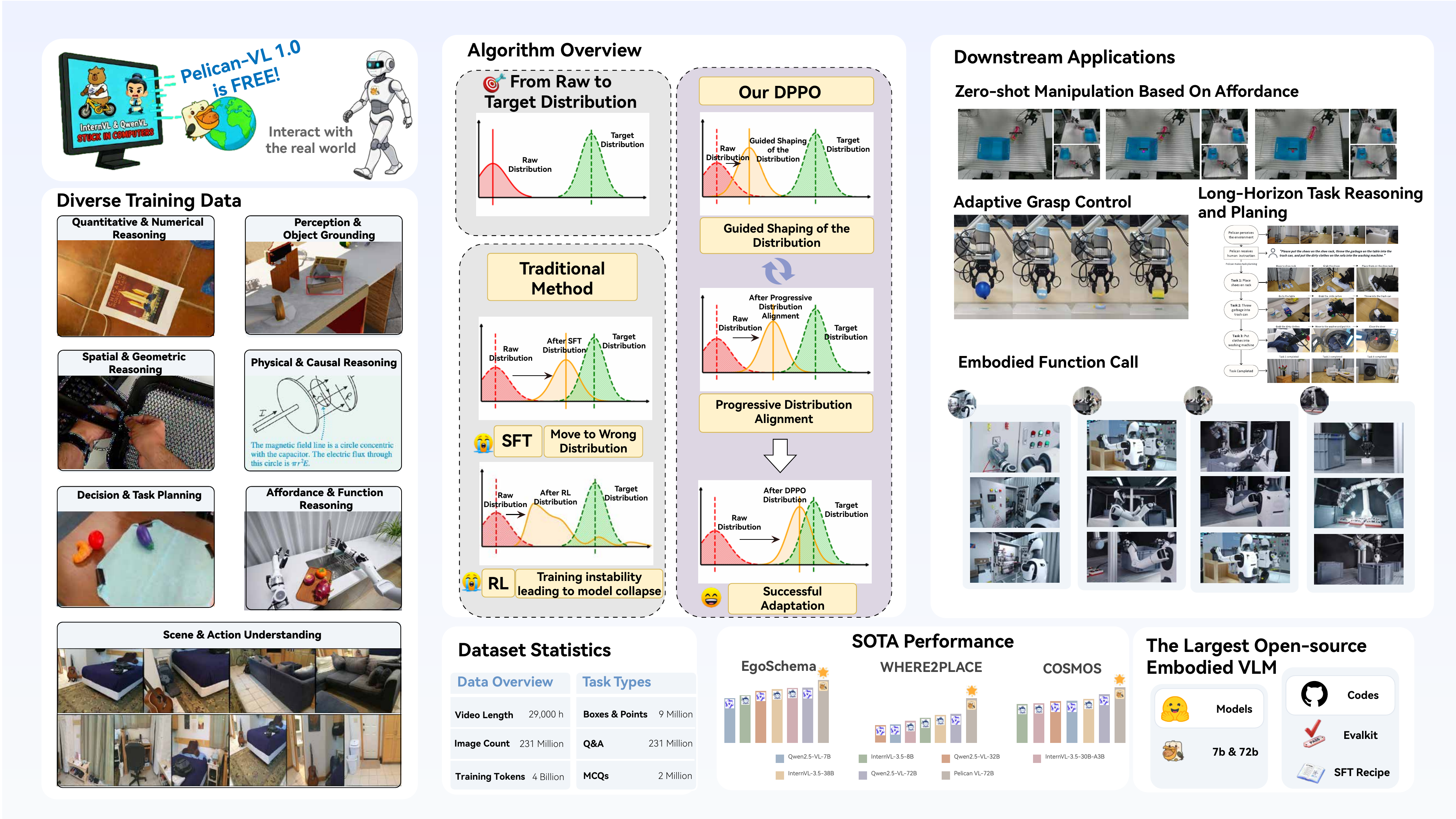}
    \label{fig:teaser}
\end{figure}

\section*{Abstract}
Recent advancements in Vision-Language Models (VLMs) have unlocked remarkable capabilities for digital-world perception. However, the translation from digital perception to embodied cognition remains a fundamental challenge. General-purpose VLMs, trained on internet-scale data, exhibit critical deficits in reasoning about complex spatial relationships, inferring temporal-causal chains, and making valid judgments about real-world interactive properties.

This report presents \model, a new family of open-source embodied brain models with parameter scales ranging from 7 billion to 72 billion. Our explicit mission is clearly stated as: To embed powerful intelligence into various embodiments. \model is currently the largest-scale open-source embodied multimodal brain model. Its core advantage lies in the in-depth integration of data power and intelligent adaptive learning mechanisms. Specifically, metaloop distilled a high-quality dataset from a raw dataset containing 4+ billion tokens. \model is trained on a large-scale cluster of 1000+ A800 GPUs, consuming over 50k+ A800 GPU-hours per checkpoint. This translates to a 20.3\% performance uplift from its base model and outperforms 100B-level open-source counterparts by 10.6\%, placing it on par with leading proprietary systems on well-known embodied benchmarks.

We establish a novel framework, DPPO (Deliberate Practice Policy Optimization), inspired by human metacognition to train Pelican-VL 1.0. We operationalize this as a metaloop that teaches the AI to practice deliberately, which is a RL-Refine → Diagnose → SFT loop. 
This loop employs Reinforcement Learning (RL) in two key roles: (1) for \textbf{skill refinement} (aligning the reference policy with target policy), and (2) for \textbf{autonomous weakness detection} (via rollouts). The hard cases discovered from these rollouts are then refined for Supervised Fine-Tuning (SFT). This SFT process facilitates \textbf{embodied competence expansion} (widening the policy's distribution).
Furthermore, we provide a theoretical interpretation of the cyclic procedure from the perspective of unified preference learning. Our AI infrastructure uniquely enables our framework to support 72B-scale, mixed-modal RL training with long-video data.

We validate \model through extensive real-world experiments on tasks including: (1) contact-rich tactile manipulation, where it is the first VLM to close the sensorimotor loop by predicting and continuously refining grasp force, (2) task-oriented affordance reasoning for pick-and-place, and (3) industry-first long-horizon planning achieved by a multi-agent system with a unified one brain controlling diverse robotic platforms. We are open-sourcing inference codebase and 7B\&72B base model checkpoints. We hope to empower the community to train and customize their own embodied brain models.

\setcounter{tocdepth}{2}
\pagebreak
 \begingroup
   \hypersetup{hidelinks}
   \tableofcontents
 \endgroup
\pagebreak

\begin{figure}[h] 
    \centering
    \begin{subfigure}[t]{0.5\textwidth}
        \centering
        \includegraphics[width=1.0\textwidth]{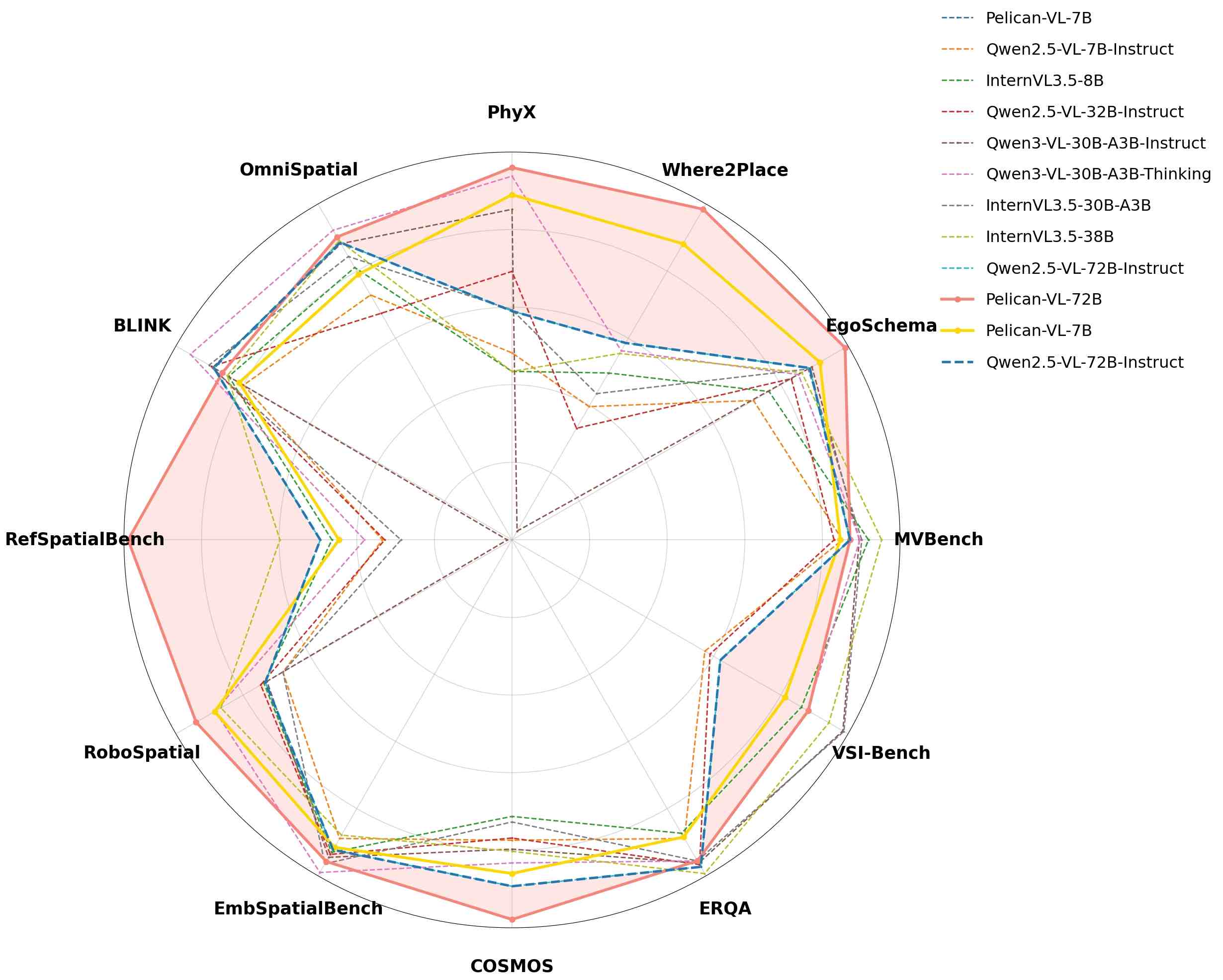}
    \end{subfigure}%
    \begin{subfigure}[t]{0.5\textwidth}
        \centering
        \includegraphics[width=1.0\textwidth]{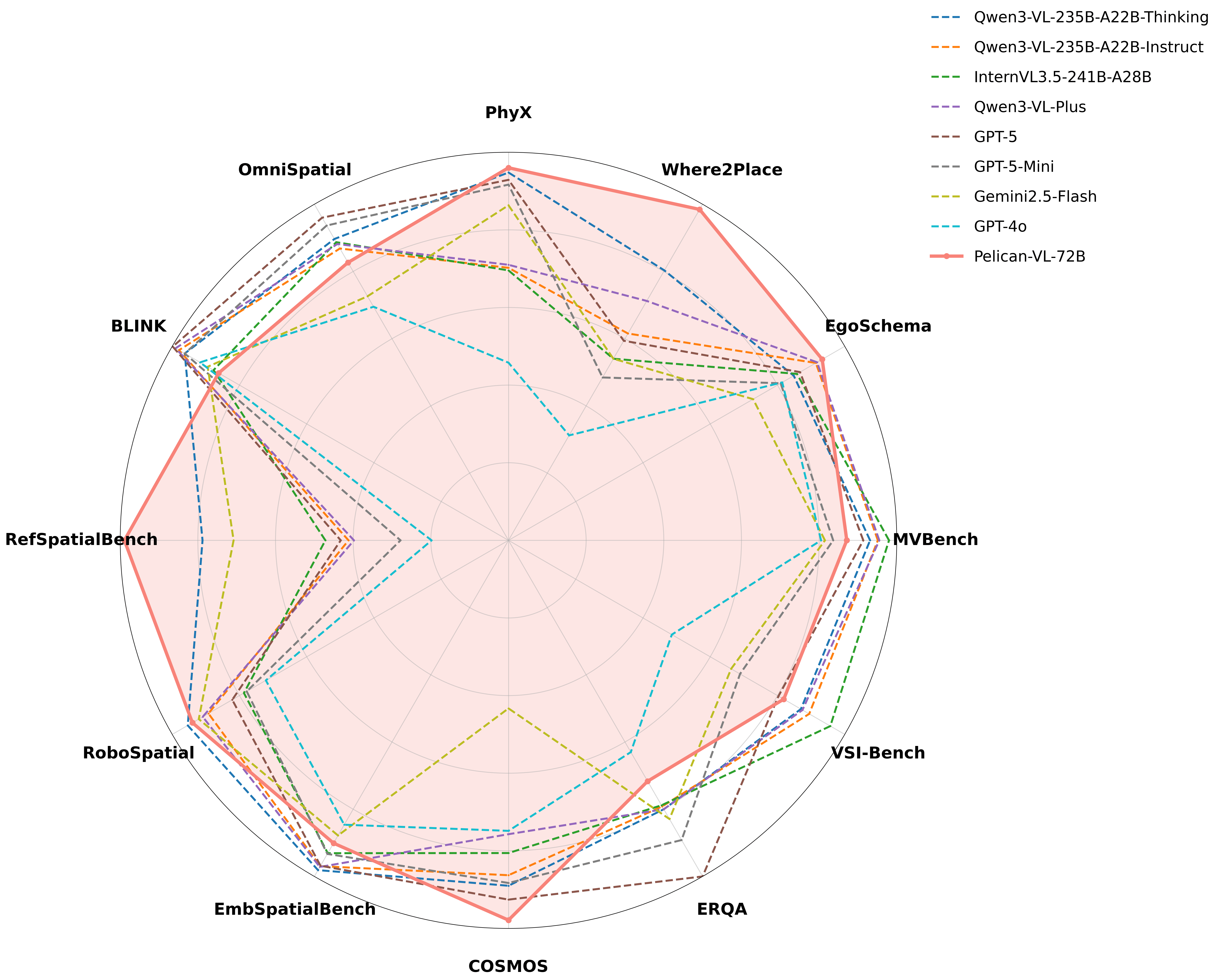}
    \end{subfigure}
    
    \caption{
        Performance comparison of Pelican-VL1.0. 
        \textbf{(Left)} Comparison against models with $\le$100B parameters. The shaded (pink) region highlights the performance gain over our baseline. 
        \textbf{(Right)} Comparison against models with $>$100B parameters, including leading open-source and proprietary models, where our model also demonstrates SOTA performance.
        }
\end{figure}

\section{Introduction}
The pursuit of Artificial General Intelligence (AGI) represents a foundational aspiration in science. A critical milestone on this quest is the creation of a universal and versatile embodied brain, a machine that can truly perceive, reason, and interact with the physical world~\cite{khan2025foundation,xiang2025parallels}. Recent advancements in Vision Language Models (VLMs) have unlocked remarkable capabilities for digital world perception. However, one possible route of general intelligence requires grounding these models in physical reality, a translation that reveals a fundamental barrier. General purpose VLMs, trained on passive, disembodied internet scale data, exhibit critical deficits across the core pillars of embodied intelligence. They often struggle with robust spatial understanding, long horizon temporal reasoning, and intuitive physical causality. This is not a simple knowledge gap but a fundamental limitation that proves insufficient for the robust, real world interaction that AGI demands.

In response, the field has bifurcated into two main strategies. The first is data scaling, where efforts like Google's Gemini Robotics~\cite{team2025gemini}, $\pi_{0.5}$~\cite{intelligence2504pi0}, GR00T N1~\cite{bjorck2025gr00t}, and GR3~\cite{cheang2025gr}, demonstrate high performance by adapting massive foundation models through data pyramid scaling~\cite{brohan2023rt,driess2023palm,ahn2022can}. The second is architectural refinement, with models like Helix~\cite{figure2025helix} and Wall OSS~\cite{zhai2025igniting}. This approach leverages a hierarchical architecture, using a large VLM for high level reasoning and a separate, smaller, reactive policy for action generation. While both strategies are foundational, they remain incomplete. The data scaling paradigm, while a necessary brute force strategy, often lacks a simple yet effective framework to manage and refine learning from such vast, heterogeneous data. Conversely, the architectural paradigm introduces vital, layered reasoning but often operates on smaller, specialized datasets, lacking the sheer scale required for true generalization. The critical missing piece is a unified framework that combines the power of massive scale data with an intelligent, adaptive learning mechanism. Unlike internet scale text, high quality embodied data is inherently scarce, expensive, and difficult to collect. 

Therefore, to unify these two paradigms massive scale and intelligent architecture we introduce \model, a new family of open source embodied brains (scaling parameters from 7B to 72B). Our mission is a clear declaration: to embed powerful intelligence in every embodiment.Grounded in the brute force paradigm, \model demonstrates that massive scale is a prerequisite, which is then harnessed by our new training framework: Deliberate Practice Policy Optimization (DPPO). This DPPO framework is inspired by a uniquely human capability, computational metacognition, or learning how to learn. This approach embraces the challenge of processing massive, petabyte scale data, actively maximizing the utility of every interaction, and enabling the creation of robust, generalizable skills that are only possible with vast, comprehensive resources.

We operationalize this DPPO framework through a metaloop that teaches the AI to practice deliberately. Theoretically, this entire process is grounded in a unified preference learning framework that leverages a powerful, synergistic SFT-RL relationship. From a distributional perspective, the loop uses Reinforcement Learning (RL)—driven by a data screener reward model—for two roles: (1) skill refinement (i.e., locally adjusting the policy's probability distribution $P_{\theta}$) and (2) autonomous weakness detection (i.e., discovering hard cases, which are distant distribution modes where $P_{\theta}$ has near-zero probability). Because RL (a local optimization) struggles to jump to these distant modes, the loop then systematically distills these discovered hard cases into new, correct data for Supervised Fine-Tuning (SFT). This SFT phase performs embodied competence expansion by globally pulling the policy's distribution to connect with these newly discovered, distant target distributions. We name this entire process the RL-Refine $\rightarrow$ Discover $\rightarrow$ SFT framework. Furthermore, when this loop detects performance saturation, it identifies the deficit as conceptual and strategically infuses curated general-domain knowledge (also via SFT) to un-stuck the learning process.

This massive investment in data and computing establishes a new foundation for capability. Crucially, simply scaling compute is insufficient, often leading to instability or catastrophic forgetting. We conducted extensive real world experiments to systematically evaluate the unique effectiveness of our DPPO framework in harnessing this massive scale. These foundational improvements are the direct drivers of its benchmark success: our framework not only demonstrates the stability to absorb this 50k+ GPU hours training budget for per checkpoint but also successfully translates this scale into tangible accuracy gains of 25.7\% increase in spatial understanding (i.e., Spatial-Physical) and 15.1\% increase in temporal reasoning (i.e., CoT-Reasoning) compared with the backbone model.

We validated Pelican-VL 1.0 through extensive real-world experiments on tasks including: (1) contact-rich tactile manipulation, where it is the first VLM to close the sensorimotor loop by predicting and continuously refining grasp force, (2) task-oriented affordance reasoning for pick-and-place, and (3) industry-first long-horizon planning achieved by a multi-agent system with a unified one brain controlling diverse robotic platforms. We open-source the inference codebase, the SFT and LoRA training codes, and the 7B \& 72B base model checkpoints. The notable contributions of \model are summarized as follows:

\begin{itemize}
    \item \textbf{Open-Source Embodied Brain with SOTA performance and Toolchain.} We release \textbf{\model}, a family of embodied brain models (7B–72B) achieving state-of-the-art performance and rivaling proprietary systems. Together with the complete \textbf{DPPO} toolchain, this open release provides a unified foundation for advancing embodied AI research.

    \item \textbf{A novel training framework-Deliberate Practice Policy Optimization (DPPO).} We introduce \textbf{DPPO}, a metacognitive training framework implemented via the \textbf{Metaloop} engine. It unifies RL and SFT in RL-Refine → Diagnose → SFT loop for targeted skill refinement and competence expansion.

    \item \textbf{Scalable Data Curation via Metaloop Selection.} The Metaloop also drives data curation by identifying skill gaps and distilling hard cases from large-scale trajectories, yielding efficient, high-quality datasets and a scalable playbook for data-intensive embodied learning.
\end{itemize}

\begin{figure}
    \centering
    \includegraphics[width=0.95\linewidth]{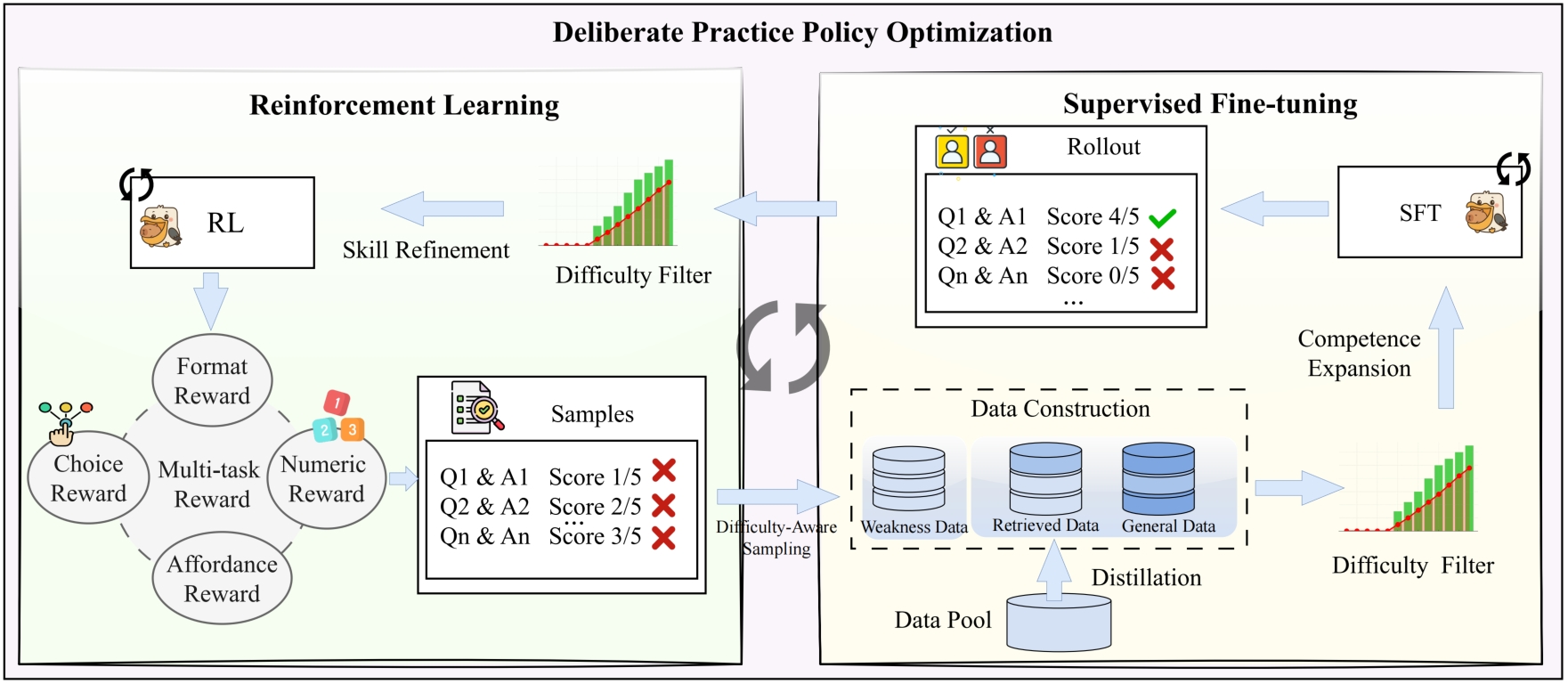}
    \caption{Overview of our training framework. This framework implements an iterative RL-SFT loop that leverages Rollout Logging and Difficulty-Aware Sampling to dynamically curate data. This adaptive data selection process is designed to achieve two complementary objectives: rapid capability enhancement during the RL phase and stable modal alignment during the SFT phase.}
    \label{fig:weakness}
\end{figure}

\section{Methodology}
In this section, we first define the embodied problem in Section~\ref{sec:problem_formulation}, outlining the composite multimodal scenarios and key embodied tasks. We then introduce the DPPO framework in Section~\ref{sec:dppo}, a metaloop engine designed to manage the computationally intensive multimodal complexity of our approach~\cite{flavell1979metacognition,brown1987metacognition}. In DPPO, we implement a self-evolving RL–SFT scheme to iteratively refine embodied capabilities and break through the model’s performance ceiling. Finally, we establish the theoretical foundations of DPPO in Section~\ref{sec:expl}, showing how SFT and RL unify under a coherent preference-learning paradigm.

\subsection{Problem Formulation: Embodied Scenarios and Tasks} \label{sec:problem_formulation}
Embodied intelligence entails aligning digital world with the physical world via multi-modal perception, interaction, and reasoning ~\cite{liu2024embodied}. We extend this perspective to formulate our embodied scenario as a composite multimodal environment that integrates perception, scene understanding, task planning, physical manipulation, and human interaction. For a given multimodal input $(x_v, x_t)$,  where $x_v$ is the visual input (images or video frames), $x_t$ is the textual input (instructions, questions, or contextual descriptions), the model predicts the output $\hat{y}$ through a parameterized mapping:

\begin{equation}
\hat{y} = f_\theta(x_v, x_t)
\end{equation}
where $\hat{y}$ denotes the target output, which can be reasoning traces, task plans, action sequences, or structured function calls.

During training, the model parameters $\theta$ are optimized to minimize the discrepancy between the output $\hat{y}$ and the ground truth $y$ using a task-appropriate loss function:

\begin{equation}
\mathcal{L}(\theta) = \mathbb{E}_{(x_v, x_t, y)} \Big[ \ell(f_\theta(x_v, x_t), y) \Big]
\end{equation}
where $\ell$ denotes the loss function. 

This unified formulation involves a high-dimensional state-action space and complex multi-task demands (e.g., causal-temporal inference, affordance reasoning, and function calls), making single-stage training (pure SFT or RL) impractical. The search space is too vast for RL to explore efficiently, while expert data remains too sparse for SFT to generalize reliably.

\subsection{Training Framework: Deliberate Practice Policy Optimization}\label{sec:dppo}

Inspired by the cognitive science principle of deliberate practice, we propose DPPO that operationalizes this principle through a dynamically regulated metaloop.  Rather than a monolithic optimization pipeline, the metaloop alternates between two synergistic phases: a RL phase that enhances weak or brittle abilities through exploratory interaction, and a SFT phase that consolidates and generalizes the knowledge gained through exploration. Together, these alternating phases form a self-improving loop that enables the model to continually diagnose, target, and remediate embodied weaknesses by effectively harnessing and refining knowledge from large-scale, heterogeneous data.

Formally, let $\theta$ denote the model parameters and $k$ index the metaloop iterations. At each iteration, a phase selector $\sigma_k \in \{\textsc{\textit{RL}},\,\textsc{\textit{SFT}}\}$ determines which objective is active, ensuring that the model optimizes only one phase-specific loss at a time:
\begin{equation}
\label{eq:phase-objective}
\mathcal{L}_{\sigma_k}(\theta) =
\begin{cases}
\mathcal{L}_{\textsc{GRPO}}(\theta; \pi_{\mathrm{ref}}), & \text{if } \sigma_k=\textsc{\textit{RL}}\\[4pt]
\mathcal{L}_{\textsc{SFT}}(\theta; \mathcal{D}^{\,k+\tfrac{1}{2}}_{\textsc{RL}} \cup \mathcal{D}_{\text{General}}^k) & \text{if } \sigma_k=\textsc{\textit{SFT}}
\end{cases}
\end{equation}
where $\pi_{\mathrm{ref}}$ is the fixed reference policy used for the Kullback-Leibler (KL) regularization term in the Group Relative Policy Optimization (GRPO) objective.

Each full training cycle, $k$, consists of one RL phase followed by one SFT phase. The SFT data, $\mathcal{D}^{\,k+\tfrac{1}{2}}_{\textsc{RL}} \cup \mathcal{D}_{\text{General}}^k$, is a union of two complementary \emph{dynamic} data sources:
\begin{enumerate}
    \item The term $\mathcal{D}^{\,k+\tfrac{1}{2}}_{\textsc{RL}}$: The dataset of trajectories \textbf{collected} from the RL phase just completed within the \emph{same} $k$-th cycle.
    \item The term $\mathcal{D}_{\text{General}}^k$: A new dataset of instructions \textbf{dynamically generated} (e.g., synthetically) for the $k$-th cycle. This generation process is guided by the model's current abilities and the weaknesses identified from the $\mathcal{D}^{\,k+\tfrac{1}{2}}_{\textsc{RL}}$ data.
\end{enumerate}
This design establishes a tight, iterative loop. The SFT phase fine-tunes the model on two synergistic data sources: (1) the raw trajectories from the recent RL phase and (2) a set of new instructions synthesized to specifically target the abilities the model struggled with. This ensures a focused refinement on its most critical weaknesses.

During each iteration, the active objective governs parameter updates according to its own learning dynamics:
\begin{equation}
\label{eq:phase-update}
\theta_{k+1} = \theta_k - \eta_{\sigma_k}\,\nabla_\theta \mathcal{L}_{\sigma_k}(\theta_k)
\end{equation}
where $\eta_{\sigma_k}$ denotes the learning rate associated with the current phase.  
Intuitively, the RL phase drives exploration and targeted skill enhancement, while the SFT phase absorbs and stabilizes these improvements within a supervised objective.

The metaloop process can be expressed as an alternating composition of phase-specific update operators:
\begin{equation}
\theta_{k+\tfrac{1}{2}} =
\underbrace{\mathcal{U}_{\textsc{RL}}(\theta_k)}_{\text{Exploratory Grounding (GRPO})}
\quad\text{\Large$\circlearrowright$}\quad
\theta_{k+1} =
\underbrace{\mathcal{U}_{\textsc{SFT}}(\theta_{k+\tfrac{1}{2}};\,\mathcal{D}^{\,k+\tfrac{1}{2}}_{\textsc{RL}} \cup \mathcal{D}_{\text{General}}^k)}_{\text{Targeted Remediation (SFT on }\mathcal{D}^{\,k+\tfrac{1}{2}}_{\textsc{RL}} \cup \mathcal{D}_{\text{Gen}}^k)}.
\end{equation}
Here, $\mathcal{U}$ denotes the \textbf{update operator} for a given training phase, which transforms the model parameters $\theta$ from one state to the next. The equation illustrates the two-stage process within a single cycle $k$: first, the \textbf{Exploratory Grounding} operator, $\mathcal{U}_{\textsc{RL}}$, applies the GRPO algorithm to update the model parameters from $\theta_k$ to $\theta_{k+\tfrac{1}{2}}$. Second, the \textbf{Targeted Remediation} operator, $\mathcal{U}_{\textsc{SFT}}$, takes these intermediate parameters and applies SFT using the mixed dataset ($\mathcal{D}^{\,k+\tfrac{1}{2}}_{\textsc{RL}} \cup \mathcal{D}_{\text{General}}^k$) to produce the final parameters $\theta_{k+1}$ for that cycle. This operator form highlights that Metaloop alternates deliberately between two mutually exclusive learning modes—exploration through RL and consolidation through SFT. Metaloop consists of two key components: an Exploratory Grounding, which uses RL to probe and strengthen weak embodied capabilities; and a Targeted Remediation, which applies SFT to absorb and stabilize the knowledge acquired during exploration. These complementary loops form a unified and generalizable paradigm for continuously improving embodied competence.

\subsubsection{Exploratory Grounding}

Unlike conventional RL, which focuses purely on maximizing cumulative rewards, this phase extends the function of RL from simple reward maximization to exploratory grounding.We adopt a GRPO-based reinforcement learning paradigm, equipped with multi-modal, multi-task reward functions .

Formally, the policy gradient is expressed as:
\begin{equation}
  \nabla_\theta \mathcal{L}_{GRPO} = \mathbb{E}_{(x,y)\sim\pi_{ref}}[w(x,y) \nabla_\theta \log \pi_\theta(y|x)]
\end{equation}
where \(w(x, y)\) denotes the normalized reward weight derived from the rule-based scoring function that compares the current policy \(\pi_\theta\) against a reference policy \(\pi_{\text{ref}}\).

\begin{table}
\centering
\small
\setlength{\tabcolsep}{6pt}
\renewcommand{\arraystretch}{1.35}
\caption{Rule-based multi-task reward design.}
\begin{tabular}{>{\centering\arraybackslash}m{3cm} | >{\centering\arraybackslash}m{3cm} | >{\centering\arraybackslash}m{9.5cm}}
\toprule
\textbf{Task Type} & \textbf{Reward Type} & \textbf{Description} \\
\midrule
Affordance Reasoning &
Affordance Reward &
Evaluates whether predicted manipulations are physically feasible and contextually valid, including grasping and placement correctness. \\
\hline
Counting and Distance Estimation &
Numeric Reward &
Rewards accurate quantitative perception and spatial measurement across objects and scenes. \\
\hline
Causal and Temporal Reasoning &
\multirow{4}{*}{Choice Reward} &
Assesses logical and temporal consistency in sequential task execution and causal dependencies. \\
\cline{1-1}\cline{3-3}
Task Success Evaluation & &
Determines whether the trajectory fulfills task-specific completion rules, such as target achievement or valid function-call execution. \\
\cline{1-1}\cline{3-3}
Task Planning & &
Rewards coherent hierarchical action decomposition that aligns sub-goals with overall objectives. \\
\cline{1-1}\cline{3-3}
Task Prediction & &
Measures accuracy in predicting task intents, expected outcomes, or next-step actions from multimodal inputs. \\
\bottomrule
\end{tabular}
\label{tab:reward_design}
\end{table}

\vspace{4pt}
\noindent
\textbf{Multi-Modal and Multi-Task Reward.}
To promote broad embodied competence, we construct a rule-based multi-task reward function covering six core objectives: affordance reasoning, counting and distance estimation, causal and temporal reasoning, task success evaluation, task planning, and task prediction. These objectives jointly guide the model to balance perception, reasoning, and planning abilities across multimodal embodied tasks. The detailed reward definitions are summarized in Table~\ref{tab:reward_design}. In addition, to stabilize both training and evaluation, we employ a format reward to constrain the model’s output. The format reward is computed by verifying whether the generated response strictly adheres to the required structure—specifically, that it contains a step-by-step reasoning trace and a correct final answer.

Each rollout \(\tau = (s_t, a_t, r_t)\) receives a composite reward that combines a general format correctness term with a task-specific reward:
\begin{equation}
R(\tau) = \lambda_{\text{fmt}} R_{\text{fmt}}(\tau) + \lambda_{\text{task}} R_{\text{task}}(\tau)
\end{equation}
where \(R_{\text{fmt}}(\tau)\) measures the structural validity of the model’s output, and \(R_{\text{task}}(\tau)\) reflects the rule-based reward associated with the specific embodied objective, such as affordance reasoning or task planning.  

\vspace{4pt}
\noindent
\textbf{Rollout Logging and Difficulty-Aware Sampling.}
During GRPO training, all rollout trajectories are automatically logged and analyzed to construct a \textit{difficulty-aware data buffer}. For each trajectory, we compute a difficulty score based on both reward success rate:
\begin{equation}
D(\tau) = 1 - \text{SuccessRate}(\tau)
\end{equation}

where $\text{SuccessRate}(\tau)$ is defined as the pass@k probability (i.e., the probability of at least $k$ successful outcomes out of $n$ rollouts) for task $\tau$ during the training rollout. A higher \(D(\tau)\) thus indicates greater task uncertainty or inconsistent success.
Samples with the highest difficulty scores, representing weakly mastered or ambiguous abilities, are automatically prioritized for inclusion in the subsequent \textbf{SFT} phase. This adaptive sampling procedure ensures that SFT data directly target the model's most critical weaknesses, forming a closed RL-SFT loop consistent with the principle of Deliberate Practice, which emphasizes focused improvement through iterative feedback and refinement.

\paragraph{RL training and stopping criterion.}
Throughout RL, we monitor the \emph{Task Saturation} indicator $\mathrm{TS}(t)$, which quantifies the proportion of tasks that are no longer yielding efficient learning. Formally, the overall saturation across all task groups $\mathcal{T}$ is computed as:
\begin{equation}
\mathrm{TS}(t) = \frac{1}{|\mathcal{T}|}\sum_{i \in \mathcal{T}}\mathrm{TS}_i(t)
\end{equation}
where $\mathrm{TS}_i(t) \in [0, 1]$ is the individual saturation score for task group $i$. This score itself is a benchmark quantifying the degree of saturation, calculated from factors such as: (1) \textbf{Degenerate Rollouts}, the proportion of validation rollouts for the task that are identical (i.e., 100\% success or 100\% failure), and (2) \textbf{Stagnated Progress}, the proportion of recent training steps (e.g., 20k steps) during which the task's success rate has not changed.

The RL stage is automatically terminated when the \emph{average} saturation score $\mathrm{TS}(t) \ge 0.7$. This adaptive rule ensures that each RL round halts once learning efficiency plateaus, preventing unnecessary computation and overfitting.

\subsubsection{Targeted Remediation}

This phase corresponds to the SFT component of DPPO and serves as the counterpart to the exploratory stage. The targeted remediation phase aims to consolidate and generalize the knowledge discovered during RL.  
It achieves this by re-anchoring the policy through high-quality data refinement and structured knowledge infusion.

\paragraph{Data Construction and Knowledge Infusion.}  
Following each RL round, all rollout trajectories are analyzed to identify both improved behaviors and unresolved weaknesses. Rather than relying solely on successful rollouts, we explicitly target the model’s unmastered abilities to construct a new supervision corpus for fine-tuning.  
The enhanced dataset \(\mathcal{D}_{\text{SFT}}\) is composed of three complementary sources:
\begin{equation}
\mathcal{D}_{\text{SFT}} = \mathcal{D}_{\text{weak}} \cup \mathcal{D}_{\text{assoc}} \cup \mathcal{D}_{\text{gen}}
\end{equation}
where \(\mathcal{D}_{\text{weak}}\) consists of hard or incorrectly solved rollout samples identified by the Difficulty-Aware Sampling mechanism in the RL phase, \(\mathcal{D}_{\text{assoc}}\) includes related embodied samples retrieved from the dataset according to the weak ability dimensions, and \(\mathcal{D}_{\text{gen}}\) contains additional data generated by vision-language models to enrich contextual and linguistic diversity. Together, these components enable the SFT phase to perform targeted knowledge infusion—transforming the model’s observed weaknesses into structured supervision signals for continual policy reinforcement.

\paragraph{Supervised Consolidation.} After the construction of \(\mathcal{D}_{\text{SFT}}\), the model undergoes supervised fine-tuning to consolidate the skills discovered during reinforcement learning. By globally integrating localized RL improvements into the policy distribution, the model mitigates catastrophic forgetting and achieves stronger generalization across diverse embodied domains.

Formally, the SFT objective follows the standard maximum likelihood estimation (MLE) formulation:
\begin{equation}
\mathcal{L}_{SFT}(\theta) = -\mathbb{E}_{(x,y)\sim D_{SFT}}\left[ \sum_{(x,y) \in \tau^*} \log \pi_\theta(y|x) \right]
\label{eq:sft_loss}
\end{equation}
where \((x, y)\) denotes a multimodal input–output pair sampled from the enhanced dataset \(\mathcal{D}_{\text{SFT}}\), 
and \(\pi_\theta\) represents the policy parameterized by \(\theta\).  
By maximizing the likelihood of high-quality demonstrations, the model internalizes reliable action patterns and consolidates the improvements discovered during RL exploration. 

\subsection{An Explanation of DPPO: Unified Preference Learning}\label{sec:expl}
The theoretical underpinning of our framework is the unification of SFT and RL into a single, cohesive paradigm of Preference Learning (PL). We posit that these seemingly disparate training methodologies can be viewed as specific instantiations of a single, universal objective: maximizing the log-likelihood of observed preference data under a policy-conditioned probabilistic model.

\paragraph{The Unified Objective Function.} Let $\pi_\theta$ be a policy parameterized by $\theta$. We define a general preference dataset $D_{pref}$ composed of preference samples $\{c_i\}$. Each sample $c$ represents any form of preference expression, such as an expert trajectory, a win/loss pair, or a ranked list of outcomes. The universal objective is to find the optimal policy parameters $\theta^*$ that maximize the expected log-likelihood of this preference data:
\begin{equation}
\theta^* = \arg \max_{\theta} \mathbb{E}_{c\sim D_{pref}}[\log P(c|\pi_\theta)]
\end{equation}
The core difference between various PL algorithms lies in (1) the specific structure of the preference sample $c$, and (2) the choice of the probabilistic model $P(c|\pi_\theta)$ that links the observed preference to the agent's policy. We follow recent work in assuming that preferences are governed by a latent reward function $r(\tau)$, which can be implicitly defined by the policy itself relative to a reference policy $\pi_{ref}$, such that $r(\tau) = \beta \log(\pi_\theta(\tau)/\pi_{ref}(\tau))$.

We now demonstrate that SFT and GRPO are elegant special cases of this unified objective.

\paragraph{SFT.} Preference Sample $c$: The sample $c$ is a single expert trajectory $\tau^*$, which is considered axiomatically optimal. Thus, $D_{pref} = D_{SFT} = \{\tau_i^*\}$. Probabilistic Model $P(c|\pi_\theta)$: The implicit assumption in SFT is that the probability of observing the expert's preference for $\tau^*$ is directly proportional to the policy's likelihood of generating that trajectory.Substituting this into the unified objective reveals that maximizing the log-likelihood of the preference data is equivalent to minimizing the standard negative log-likelihood SFT loss $\mathcal{L}_{SFT}(\theta)$:

\begin{equation}
\log P(\tau^* | \pi_{\theta}) = \log P(\tau^*; \theta) = \sum_{(s, a^*) \in \tau^*} \log \pi_{\theta}(a^*|s)
\end{equation}

\paragraph{Group Relative Policy Optimization.} Preference Sample $c$: The sample $c$ is a ranked list of trajectories $\{\tau_1, \tau_2, \dots, \tau_k\}$, where $\tau_i \succ \tau_{i+1}$. Probabilistic Model $P(c|\pi_\theta)$: A Plackett-Luce model is employed to define the probability of observing this specific ranking. This is the product of sequential selection probabilities:
\begin{equation}
P(c|\pi_\theta) = \prod_{i=1}^k P(\tau_i \text{ chosen from } \{\tau_i, \dots, \tau_k\}|\pi_\theta) = \prod_{i=1}^k \frac{\exp(r(\tau_i))}{\sum_{j=i}^k \exp(r(\tau_j))}
\end{equation}

Substituting the implicit reward function $r(\tau)$ and maximizing the log-probability of this expression recovers the GRPO objective. This demonstrates that GRPO naturally extends preference optimization to richer, rank-based preference signals.

\paragraph{Synergistic Roles of SFT and RL from the Unified Formulation.} This unified mathematical framework provides a principled explanation for the synergy between SFT and RL (with RL instantiated here as GRPO). \textbf{(1) SFT for Knowledge Enhancement.} As shown in its formulation, the SFT objective is optimized on the dataset $D_{SFT}$ containing only positive exemplars ($\tau^*$). Consequently, the gradient of the unified objective function is exclusively directed towards increasing the likelihood of these expert behaviors. This makes SFT a direct and stable mechanism for knowledge enhancement, rapidly grounding the policy in a region of demonstrated competence and instilling a foundational understanding of the task. \textbf{(2) RL for Weakness Detection and Refinement.} In contrast, RL-based methods like GRPO operate on preference samples $c$  that include both superior and inferior outcomes. The unified objective promotes preferred behaviors while suppressing dispreferred ones, effectively driving weakness detection and refinement. This allows the policy to learn from its own suboptimal variations and correct subtle flaws that SFT alone may overlook.

\section{Data curation}
\subsection{Data for Training}
To advance the capability of \model in perception, reasoning, and planning in real-world scenarios grounded in physical knowledge, we built a diverse data pool. This pool contains \textbf{231M} images and \textbf{29k} hours of video, encompassing \textbf{231M} open-ended question-answer (QA) pairs, \textbf{9M} grounding annotations (bounding boxes or keypoints), and \textbf{2M} multiple-choice questions (MCQs) From this curated data, we sampled \textbf{1.3M} instances for SFT stages and \textbf{0.5M} for RL, amounting to 4B training tokens,  and categorizing the data into four fundamental capability areas to specifically address challenges in embodied AI:

\begin{itemize}
    \item \textbf{Physical, Spatial and Numerical Reasoning.} This category aims to build a foundational understanding of the physical world. It includes data that assesses physics-grounded reasoning in visual scenarios (e.g., \textit{PhyX}, which contains multimodal questions across core physics domains) and the understanding of spatial configurations (e.g., \textit{VSI-bench}, with tasks on measurement estimation and spatiotemporal reasoning). We also integrate data from 2D images, 3D embodied videos, and simulated scenes to cover tasks like Reasoning QA (e.g., \textit{RefSpatial}).
    
    \item \textbf{Perception, Grounding and Multi-Object Consistency.} This category focuses on connecting language to visual elements and understanding object relationships. It leverages data that combines object and space-referenced information with VQA and detection. For instance, some datasets guide actions via image keypoint prediction from language instructions (e.g., \textit{Where2Place}), while others include referring expression tasks to ensure precise object grounding (e.g., \textit{RefSpatial}).
    
    \item \textbf{Temporal, Functional and Scene Understanding.} To enable reasoning in dynamic environments, this category incorporates complex scene understanding tasks that require integrating video, images, and language (e.g., \textit{COSMOS}). It utilizes 3D embodied video data to evaluate the model's grasp of temporal sequences, object affordances, and overall scene semantics (e.g., \textit{RefSpatial}).
    
    \item \textbf{Decision Making and Task Planning.} This category targets high-level cognitive abilities for robotics. The data is designed to support decision-making in dynamic contexts (e.g., \textit{COSMOS}) and to translate language instructions into concrete actions (e.g., \textit{Where2Place}). This includes specific downstream tasks such as determining vacant spaces for placement (e.g.,  \textit{Vacant QA}) and executing object placement plans (e.g., \textit{RefSpatial}).
\end{itemize}

\subsection{Metaloop Data Selection}

\paragraph{Learning Embodied Knowledge from the Natural World.}
\label{sec:nat_vide}
To address data scarcity in embodied AI, we explore leveraging natural world data (e.g., \textit{YouTube videos}) for model improvement. We select the SpatialVID \cite{wang2025spatialvid} dataset but discard its existing expert-derived annotations (e.g., \textit{camera pose, spatial captions}) to focus on self-evolution from raw video. We use a general-purpose VLM, Qwen3VL-Plus \cite{qwen3-vl}, to generate 24 spatial QA pairs for each of 75k videos (e.g., \textit{\textless How many green cars in this video? \textgreater}, \textit{\textless The man with the blue jacket and red shirt, who is closer to the camera?\textgreater}).
We then filter this data using InternVL3.5-38B \cite{internvl3.5}, which infers answers for each question twice. A QA pair is kept if InternVL3.5's answer matches the Qwen3VL-Plus answer. If InternVL3.5's two inferences are identical to each other but differ from the original, we replace the original answer with the new consensus answer. This refinement process yields 14k QAs.

We further enhance this dataset with the 19k QA video subset from InternSpatial \cite{deng2025internspatial}, which also consists of pure video and text instructions without box or mask annotations. The final 33k QAs are categorized into eight tasks: Object Count, Object Size, Relative Distance, Absolute Distance, Appearance Order, Room Size, Relative Direction, and Route Plan. This curated dataset is employed during the SFT stage to specifically address model weaknesses identified during RL.

\paragraph{Enhancing Embodied Abilities from Weakness Data.}
\label{weaknessdataselection}

\begin{figure}[!t]
    \centering
    \includegraphics[width=0.98\linewidth]{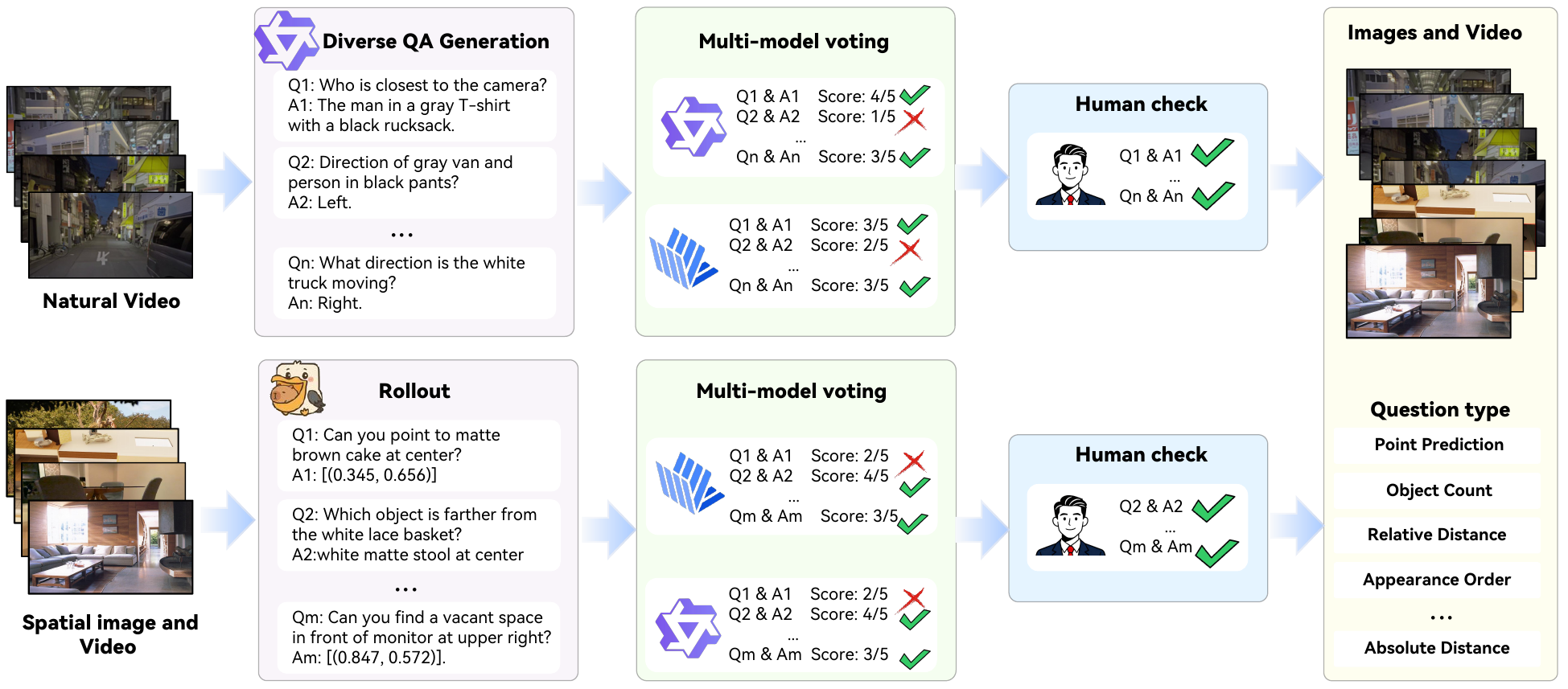}
    \caption{Overview of the metaloop data selection process.}
    \label{fig:weakness}
\end{figure}

Each loop of Metaloop requires injecting distinct weakness data tailored to the model’s current capability bottlenecks. This targeted data injection ensures that each iterative training phase addresses specific underperforming areas rather than general retraining. As shown in Figure \ref{fig:weakness}, taking the first Metaloop round as an example, after RL training we perform four rounds of rollout inference on the SFT data pool using the post-RL model. Rule-based filtering is applied to preliminarily identify weakness samples, followed by format unification. To remove low-quality or semantically ambiguous cases, the data are scored by Qwen3VL-Plus \cite{qwen3-vl} and InternVL3.5-38B \cite{internvl3.5}, and high-quality weakness samples are selected via a voting strategy. To further ensure quality, we additionally conduct random human review to guarantee the reliability of the selected weakness samples.


\section{Experiment}

In this section, we first introduce the experimental setup in Section \ref{sec:exp}. Then, we investigate the key factors influencing model performance to validate the effectiveness of DPPO. In Section \ref{sec:cyclebound}, we verify the performance change curve of Pelican after each metaloop. Section \ref{sec:finalresult} presents the results of Pelican on various benchmarks, as well as  performance on function calling in the Berkeley Function-Calling Leaderboard.

\subsection{Experimental Setup \label{sec:exp}}
We conduct three metaloop, each loop consisting of a RL phase followed by a SFT phase. In every loop, RL serves as the exploratory stage that identifies competence boundaries, while SFT consolidates and generalizes the learned behaviors to broader embodied tasks.

We adopt a curated embodied data set filtered by temporal length to ensure compact interactions focused on action. In the first loop, training data are restricted to video segments shorter than 32\,s, allowing the model to focus on short-horizon manipulation and spatial reasoning tasks. In the second loop, the temporal limit is relaxed to 64\,s, enabling the model to explore longer and more compositional trajectories as its competence grows.  

Moreover, due to variations in task difficulty and data source, the frame extraction settings differ across datasets, each video clip is sampled up to 32 frames per episode. During RL training, each rollout sequence contains up to 16 times, forming the effective temporal context for policy optimization. This progressive expansion of temporal coverage across cycles encourages the policy to generalize from short, focused tasks to more extended, temporally dependent behaviors.

\definecolor{morandiRed}{RGB}{203,160,152}
\definecolor{morandiBlue}{RGB}{173,190,202}
\definecolor{morandiBorder}{RGB}{180,180,180} 

\newcolumntype{C}[1]{>{\centering\arraybackslash}p{#1}}

\begin{figure}[htbp]
    \centering
    \includegraphics[width=\linewidth,keepaspectratio]{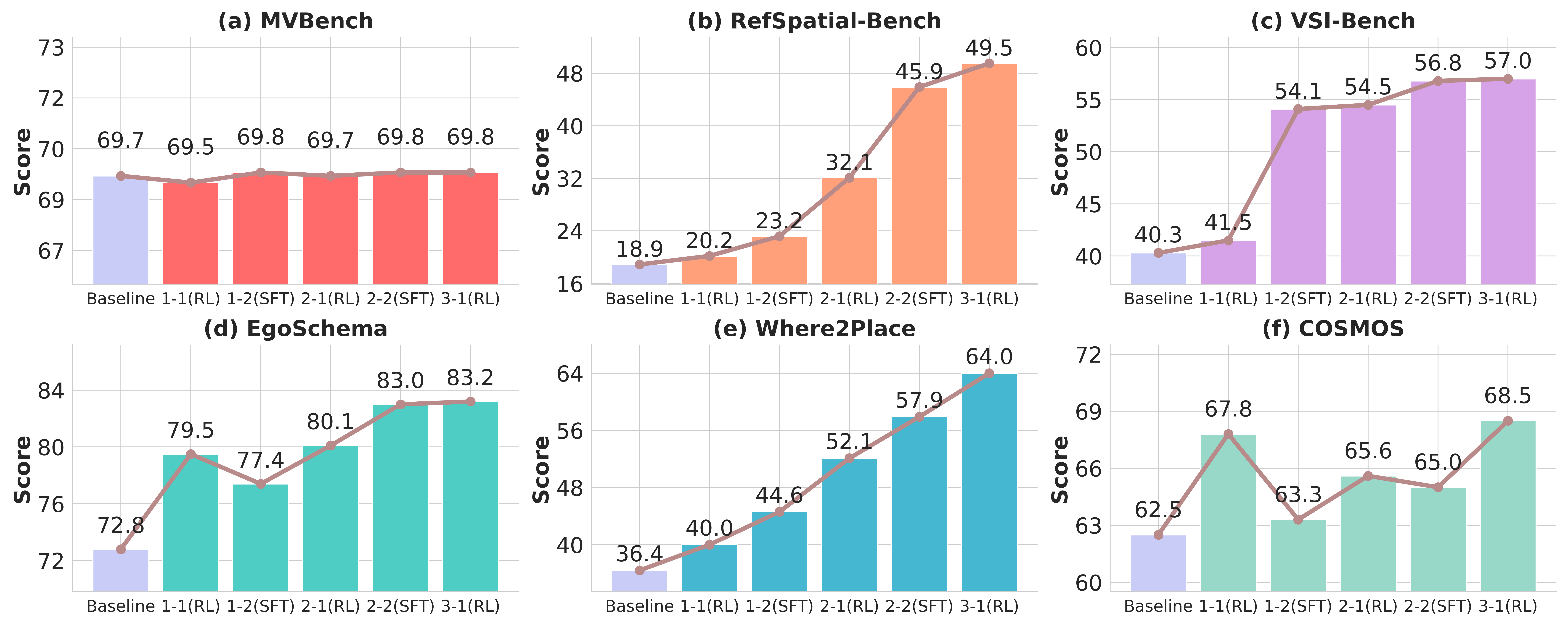}
    \caption{Performance Evolution at Each Stage of DPPO.}
    \label{fig:evolution}
\end{figure}

\begin{figure}[htbp]
\centering
\setlength{\tabcolsep}{5pt} 
\renewcommand{\arraystretch}{1.2}

\vspace{2mm} 

\begin{tabular}{ccc}
{\includegraphics[width=0.3\textwidth]{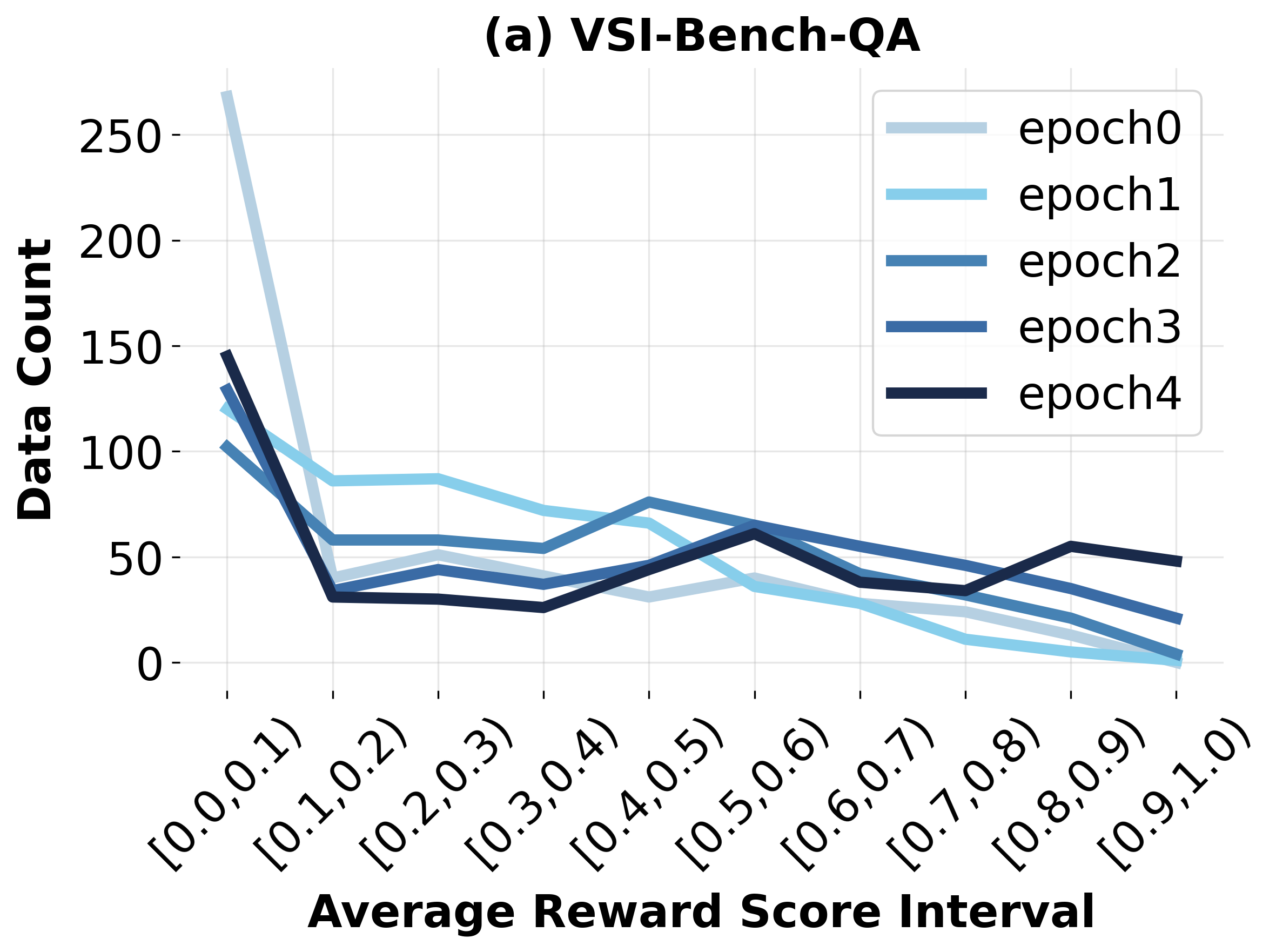}} &
{\includegraphics[width=0.3\textwidth]{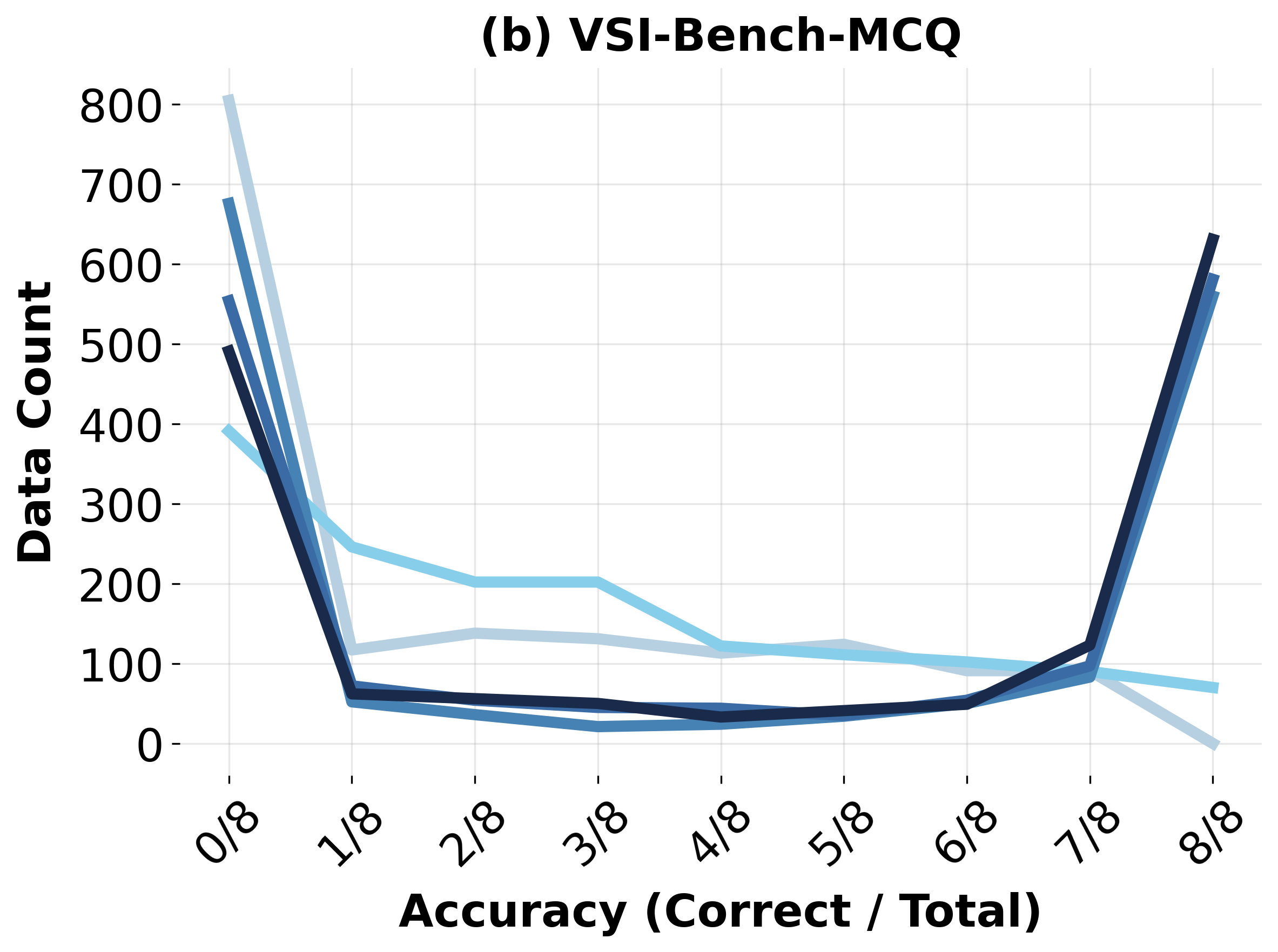}} &
{\includegraphics[width=0.3\textwidth]{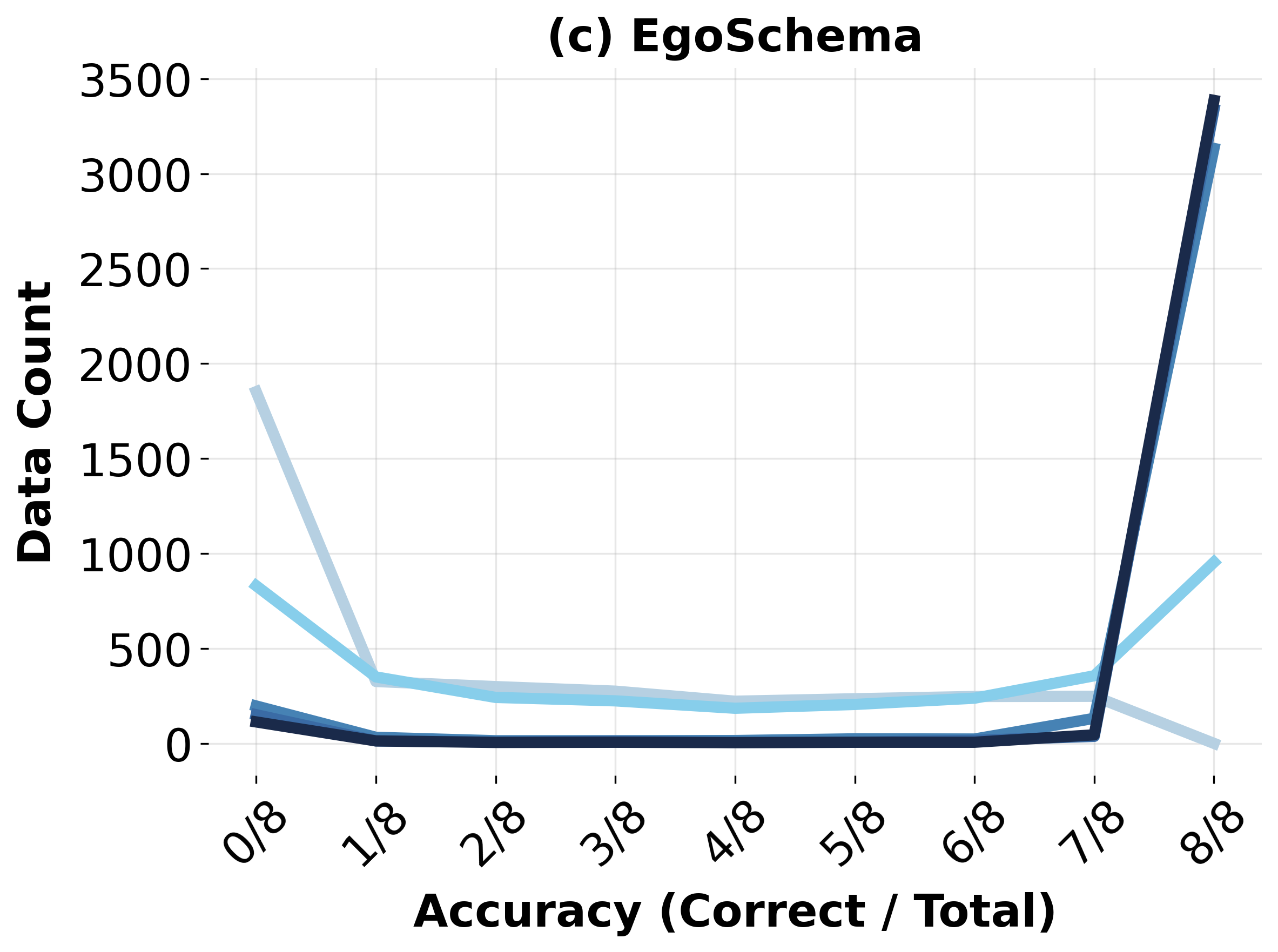}} \\
{\includegraphics[width=0.3\textwidth]{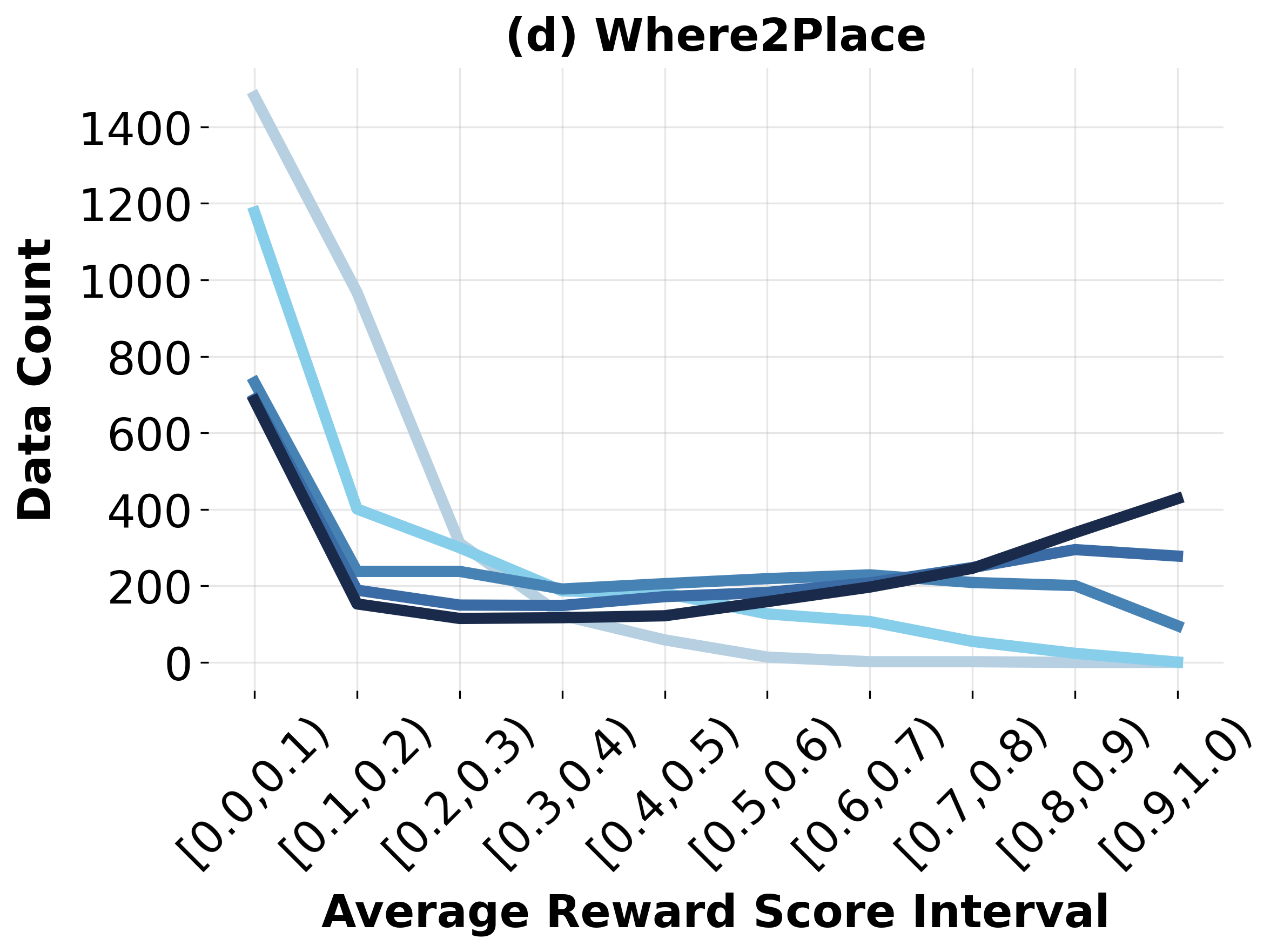}} &
{\includegraphics[width=0.3\textwidth]{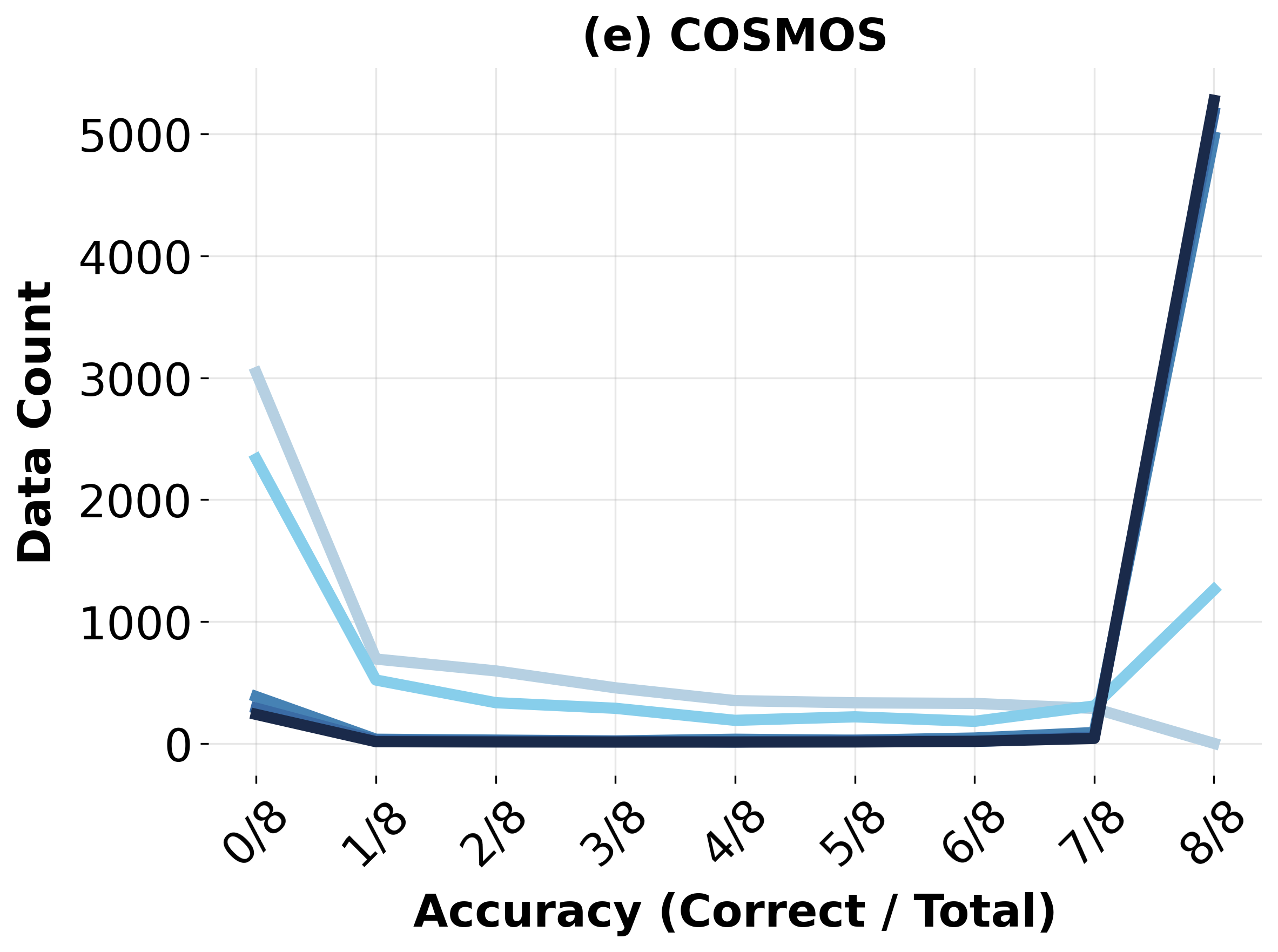}} &
{\includegraphics[width=0.3\textwidth]{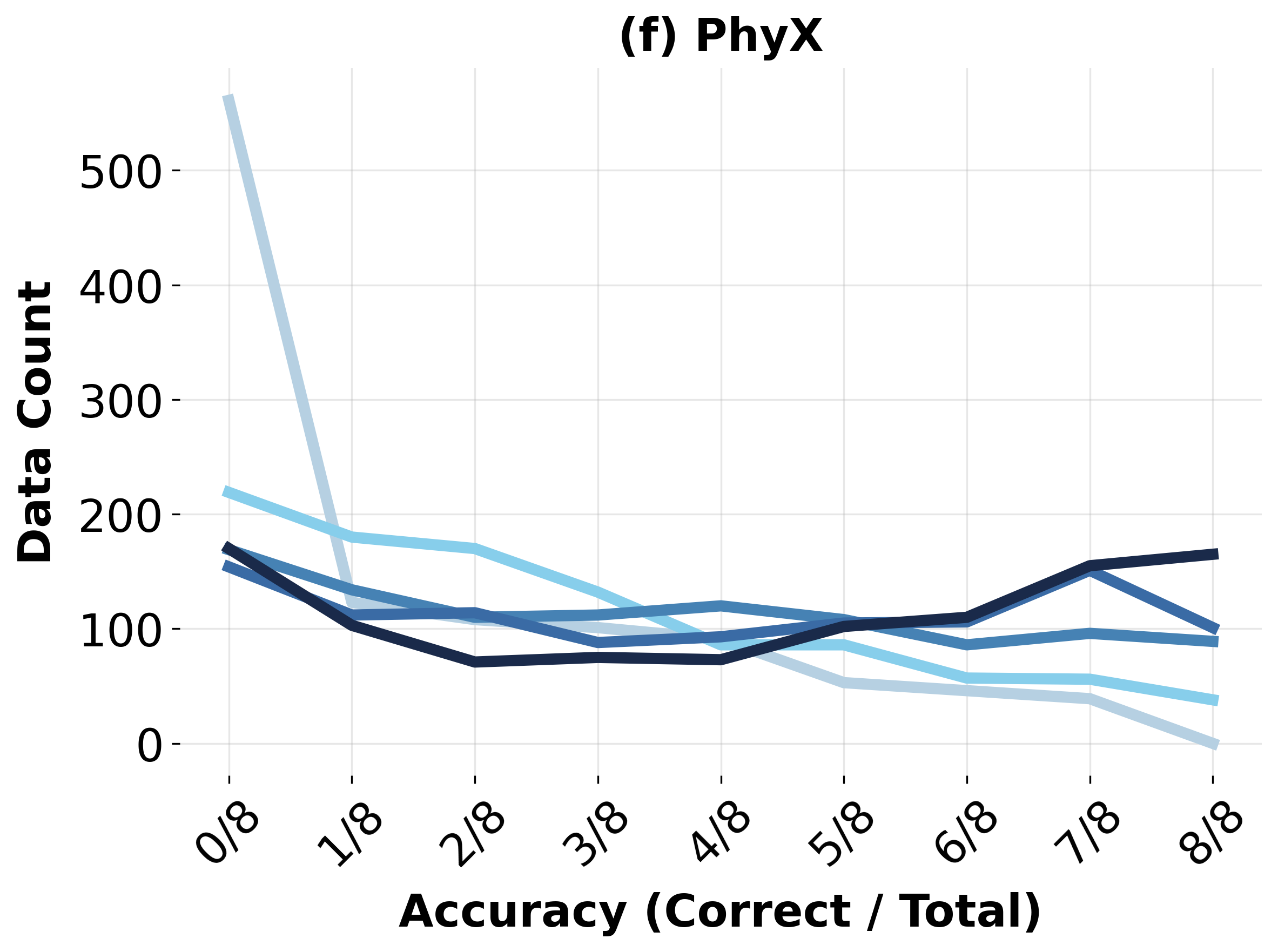}} \\
\end{tabular}

\caption{Distributional Shift of Training Data Relative to Distinct Benchmarks in RL Training. For Where2Place and VSI-Bench-QA, the rewards are numerical and the model’s performance is measured by the average score per rollout, whereas for the other datasets, the rewards are binary and performance is measured by the number of correct answers.}
\label{performance}
\end{figure}

\subsection{Performance Evolution Trajectory Across Metaloop Training \label{sec:cyclebound}}

Figure \ref{performance} shows the evolution of the 7B model's data distribution on benchmark-related tasks during the RL training process. In Figure \ref{performance} (a) and (d), the x-axis represents the average reward score. Epoch 0 indicates the scores before training, while Epoch 4 indicates the scores after four training iterations. When the scores are too high, indicating that the model has converged, RL is stopped, and the difficult samples encountered during exploration are added to the SFT training corpus, injecting the related knowledge into the model via SFT.In Figures \ref{performance} (b), (c), (e), and (f), the x-axis represents accuracy. From Figure \ref{performance} , as training progresses, RL quickly converges and discards easy samples . For samples that remain difficult, especially those with all eight rollouts incorrect, these are added to the SFT training data.

The two core components of the Metaloop: RL enhances weak or fragile embodied capabilities through exploratory interaction; SFT incorporates and consolidates the knowledge uncovered during RL exploration. As shown in Figure \ref{fig:evolution}, we discuss the performance curves of Pelican during the Metaloop process on two general benchmarks (MVBench\cite{mvbench}, EgoSchema\cite{EgoSchema}) and four embodied benchmarks (RefSpatialBench\cite{song2025robospatial}, VSI-Bench\cite{yang2024think}, Where2Place\cite{yuan2024robopoint}, COSMOS\cite{seghal2023cosmos}).

\textbf{Catastrophic Forgetting.} MVBench is a general dataset used to measure the model’s performance in general domains. The model’s performance on MVBench can indicate whether forgetting occurs during training. Figure \ref{fig:evolution} (a) shows that the model’s performance remains stable throughout training, meaning no forgetting occurs during the metaloop cyclic training. This is because when the model detects a bottleneck, it incorporates general data into metaloop for training. This data not only helps the model break through performance bottlenecks but also maintains its general capabilities, preventing catastrophic forgetting.

\textbf{Embodied abilities.} Pelican shows significant improvements on the RefSpatialBench, VSI-Bench, and Where2Place embodied datasets. At each iteration, Metaloop analyzes model diagnostics, such as rollout success rates (as shown in Figure \ref{performance}) on benchmark, to proactively schedule cycles of broad skill reinforcement through SFT and targeted weakness discovery with RL. Both learning streams follow the same probabilistic objective, enabling the training process to dynamically balance between broad knowledge reinforcement and targeted error correction. As a result, the model’s performance continuously improves throughout Metaloop training.

\subsection{Final Result}\label{sec:finalresult}

\subsubsection{Benchmark Performance}

Table \ref{tab:main_performance} presents a performance comparison between \model and other baseline models on multiple benchmarks. As shown in Table \ref{tab:100b_models}, \model achieves superior performance at the 100B-level. Compared with the Qwen2.5VL 72B-Instruct \cite{qwen2.5-vl}, \model achieves significant improvements on benchmarks in the embodied domain. This is mainly attributed to our training framework, where RL can accurately identify currently deficient capabilities, and SFT is employed to inject knowledge for weak capabilities, expanding the model's capability boundary. Furthermore, since we incorporated general data in the second round of metaloop, compared with the base model, we observe that while enhancing the model's embodied capabilities, its general capabilities do not experience a significant degradation.

\begin{table}
\centering
\caption{Overall performance comparison on benchmarks. Bold and underlined numbers indicate the best and second-best results, respectively. A dagger ($\dagger$) marks results differing from official reports or unusually low, possibly as official evaluations used model-specific prompts and the models are prompt-sensitive, while our results are obtained under a unified protocol for fair comparison. An asterisk (*) denotes results reported from official sources. Yellow cells mark our proposed \textbf{Pelican-VL 1.0} models.}
\label{tab:main_performance}

\begin{subtable}{\textwidth}
    \centering
    \caption{Performance of Models with $\leq$100B Parameters.}
    
    \label{tab:100b_models}
    \resizebox{\textwidth}{!}{%
    
    \begin{tabular}{llcccccccc >{\columncolor{pelicanlight}}c >{\columncolor{pelicanmain}}c}

    \toprule
    Category & Benchmark & Qwen2.5-VL & InternVL3.5 & Qwen2.5-VL & Qwen3-VL & Qwen3-VL & InternVL3.5 & InternVL3.5 & Qwen2.5-VL & Pelican-VL & Pelican-VL \\
    & & 7B-Instruct & 8B & 32B-Instruct & 30B-A3B-Instruct& 30B-A3B-Thinking & 30B-A3B & 38B & 72B-Instruct & 7B & 72B \\
    \midrule
    Common & MVBench & 68.1 & \underline{73.5} & 66.4 & 71.6 & 71.7 & 72.1* & \textbf{76.2} & 69.7 & 67.7 & 69.7 \\
    \midrule
    \multirow{8}{*}{Spatial-Physical} & RoboSpatial & 44.3 & 48.0 & 48.6 & 47.4$^{\dagger}$ & 57.4 & 44.3 & 56.3 & 47.7 & \underline{57.5} & \textbf{61.1} \\
    & BLINK & 55.9 & 59.0 & 62.6 & 62.3 & \textbf{67.0} & \underline{63.1} & 59.3 & 62.1 & 56.8 & 60.3 \\
    & PhyX & 43.4 & 39.1 & 62.3 & 76.7 & \underline{84.4} & 53.4 & 39.1 & 53.1 & 80.1 & \textbf{86.4} \\
    & OmniSpatial & 40.1 & 44.6 & 37.1 & 48.5 & \textbf{50.7} & 46.4 & 49.0 & 48.7 & 43.5 & \underline{49.6} \\
    & Where2Place & 25.8 & 32.3 & 21.6 & 1.7$^{\dagger}$ & 36.6 & 28.3 & 36.1 & 38.1 & \underline{57.3} & \textbf{64.0} \\
    & EgoSchema & 57.5 & 61.2 & 66.5 & 71.4 & 68.2 & 70.5 & 69.0 & 70.9 & \underline{73.3} & \textbf{79.3} \\
    & EmbSpatialBench & 71.1 & 74.4 & 74.9 & 75.6 & \textbf{79.2} & \underline{77.1} & 70.3 & 73.8 & 73.2 & 76.6 \\
    & RefSpatialBench & 16.7 & 23.2 & 16.4 & 0.6$^{\dagger}$ & 19.0$^{\dagger}$ & 14.3 & \underline{29.9} & 24.7 & 22.3 & \textbf{49.5} \\
    \midrule
    \multirow{3}{*}{CoT-Reasoning} & ERQA & 40.0 & 39.3 & 43.5 & 43.3 & 43.0 & 43.0 & \textbf{44.7} & \underline{43.8} & 39.8 & 43.0 \\
    & COSMOS & 54.2 & 49.9 & 53.8 & 55.8 & 58.3 & 50.9 & 56.2 & \underline{62.5} & 60.2 & \textbf{68.5} \\
    & VSI-Bench & 37.3 & 56.0 & 38.3 & \underline{64.0} & 57.2 & \textbf{64.2} & 61.3 & 40.3 & 52.8 & 57.3 \\
    \midrule
    Average & & 46.2 & 50.0 & 49.3 & 51.6 & \underline{57.7} & 52.3 & 53.9 & 53.0 & 57.0 & \textbf{63.8} \\
    \bottomrule
    \end{tabular}%
    }
\end{subtable}

\vspace{1cm} 

\begin{subtable}{\textwidth}
    \centering
    \caption{Performance of Models with $>$ 100B Parameters.}
    \label{tab:200b_models}
    \resizebox{\textwidth}{!}{%
    \begin{tabular}{llcccccccc>{\columncolor{pelicanmain}}c}
    \toprule
    Category & Benchmark & Qwen3-VL & Qwen3-VL & InternVL3.5 & Qwen3-VL-Plus & GPT-5 & GPT-5-Mini & Gemini2.5-Flash & GPT-4o & Pelican-VL \\
    & & 235B-A22B-Thinking & 235B-A22B-Instruct & 241B-A28B & proprietary & proprietary & proprietary & proprietary & proprietary & 72B \\
    \midrule
    Common & MVBench & 74.5 & 76.1 & \textbf{78.4} & \underline{76.4} & 73.1 & 66.9 & 65.2 & 64.4 & 69.7 \\
    \midrule
    \multirow{8}{*}{Spatial-Physical} & RoboSpatial & \textbf{62.0} & 58.0 & 51.1 & 59.1 & 53.4 & 50.6 & 59.9 & 46.9 & \underline{61.1} \\
    & BLINK & 67.4 & 68.6 & 61.4* & \underline{69.3} & \textbf{69.9} & 67.4 & 62.6 & 64.2 & 60.3 \\
    & PhyX & \underline{85.3} & 63.2 & 62.6 & 63.9 & 83.6 & 82.5 & 77.7 & 41.2 & \textbf{86.4} \\
    & OmniSpatial & 53.8 & 52.1 & 53.2 & 52.9 & \textbf{57.6} & \underline{56.2} & 43.6 & 41.7 & 49.6 \\
    & Where2Place & \underline{52.2} & 40.0 & 35.1 & 46.3 & 38.6 & 31.5 & 35.1 & 20.3 & \textbf{64.0} \\
    & EgoSchema & 72.1 & 77.7 & 72.9 & \underline{78.0} & 73.7 & 68.8 & 61.8 & 69.2 & \textbf{79.3} \\
    & EmbSpatialBench & \textbf{83.4} & 82.4 & 79.1 & \underline{82.6} & 82.3 & 79.3 & 74.6 & 71.9 & 76.6 \\
    & RefSpatialBench & \underline{39.4}$^{\dagger}$ & 20.6$^{\dagger}$ & 23.6 & 19.9 & 21.6 & 13.9 & 35.4 & 9.9 & \textbf{49.5} \\
    \midrule
    \multirow{3}{*}{CoT-Reasoning} 
    & ERQA & 48.0 & 47.5 & 47.3 & 48.0 & \textbf{60.0} & \underline{53.5} & 49.8 & 37.8 & 43.0 \\
    & COSMOS & 62.3 & 60.4 & 56.4 & 53.0 & \underline{64.8} & 61.8 & 30.3 & 52.4 & \textbf{68.5} \\
    & VSI-Bench & 60.9 & \underline{62.6}* & \textbf{67.0} & 61.2 & 56.2 & 48.2 & 46.4 & 34.0 & 57.3 \\
    \midrule
    Average & & \underline{63.4} & 59.1 & 57.4 & 59.2 & 61.2 & 56.7 & 53.5 & 46.2 & \textbf{63.8} \\
    \bottomrule
    \end{tabular}%
    }
\end{subtable}

\end{table}

Table \ref{tab:200b_models} showns that \model, trained solely on a diverse dataset consisting of 1M trajectories and 100K objects, demonstrates superior performance over even the top-performing 200B-level closed-source models, including GPT-5 and Gemini2.5-Flash, while utilizing only one-tenth of the computational resources. This further validates the effectiveness of our training strategy.

\subsubsection{Benchmark Analysis}\label{sec:bench}
Current embodied benchmarks are too coarse-grained, often providing only task-level Pass/Fail metrics. This ``black-box" evaluation is insufficient for fine-grained analysis and, critically, cannot guide iterative model updates, such as the targeted enhancement of a specific capability dimension.

To address this, we re-annotate multiple public embodied datasets according to a new taxonomy of core capability dimensions (e.g., spatial reasoning, physical causality, as detailed in Table~\ref{tab:benchmark}). This fine-grained taxonomy enables us to conduct a novel, diagnostic analysis. We use it to analyze the capability profile of various existing models, revealing systemic gaps. We also analyze our own \model's capability dimensions before and after our DPPO training, allowing us to precisely measure and demonstrate the targeted impact of our self-refine framework. 

\paragraph{Capability Taxonomy.}To systematically characterize embodied understanding, we define a unified taxonomy consisting of nine primary dimensions that span the full perceptual–cognitive–action hierarchy. 
(1) Physical \& Causal Reasoning involves predicting physical feasibility, stability, and interactions such as collisions or support relations. 
(2) Perception \& Object Grounding measures visual grounding abilities including point selection, bounding box localization, and grasp-point identification. 
(3) Quantitative \& Numerical Reasoning assesses a model’s understanding of discrete counts, continuous quantities (e.g., size, length, distance), and relative magnitudes (e.g., heavier/lighter). 
(4) Spatial \& Geometric Reasoning captures spatial layout comprehension across reference frames and topological relations such as inside/outside or adjacency. 
(5) Temporal \& Sequential Reasoning evaluates the ability to infer temporal order, state transitions, and motion dynamics across frames or events. 
(6) Affordance \& Function Reasoning focuses on understanding how objects can be manipulated, opened, poured, or used as tools under physical and kinematic constraints. 
(7) Multi-Object \& Scene Consistency examines relational coherence among multiple entities and ensures spatial–semantic consistency within complex scenes. 
(8) Scene \& Action Understanding targets high-level perception, including scene categorization, human or agent action recognition, state detection, and pose or gesture interpretation. 
(9) Decision \& Task Planning addresses goal-directed reasoning, including next-action prediction, conditional if–then logic, sequential plan generation, and sub-goal decomposition. 
Together, these nine dimensions encompass the fundamental reasoning spectrum required for real-world embodied intelligence.

\paragraph{Re-labeling Analysis.} We re-annotated 27,667 samples from ten public embodied datasets according to a unified taxonomy comprising nine primary reasoning dimensions. The overall distribution reveals a substantial imbalance between the embodied abilities. Spatial and Geometric Reasoning dominates with 8,465 samples (30.6\%), followed by Quantitative and Numerical Reasoning with 6,142 (22.2\%) and Scene and Action Understanding with 5,359 (19.4\%). In contrast, Physical and Causal Reasoning (979, 3.5\%), Affordance and Function Reasoning (632, 2. 3\%) and Decision and Task Planning (914, 3.3\%) are severely underrepresented, while Perception and Object Grounding (1,047, 3.8\%), Temporal and Sequential Reasoning (2,910, 10. 5\%), and Multiobject and Scene Consistency (1,219, 4.4\%) occupy intermediate proportions. This uneven distribution indicates that existing embodied benchmarks disproportionately emphasize spatial, numerical, and semantic reasoning, while key competencies essential for real-world robotic intelligence, such as physical feasibility, affordance perception, and long-horizon decision planning, are sparsely represented. These findings highlight the need to develop a balanced benchmark that equitably evaluates the perceptual, physical, and cognitive dimensions of embodied intelligence.

\paragraph{Limitations of Existing Embodied Benchmarks.}
As shown in Table~\ref{tab:coverage}, the comparison across ten mainstream embodied benchmarks reveals a pronounced disparity in capability coverage. Most existing datasets concentrate on a subset of embodied reasoning—particularly Spatial \& Geometric, Quantitative, and Scene \& Action Understanding—while offering limited representation of physically grounded and goal-oriented reasoning dimensions. Physical \& Causal and Affordance \& Function reasoning appear only sporadically across benchmarks, and Decision \& Task Planning is scarcely covered beyond a few simplified next-action tasks. Temporal and sequential reasoning is also weakly represented despite its importance for dynamic perception and long-horizon control. These gaps collectively indicate that current benchmarks insufficiently capture the full perceptual–cognitive–motor hierarchy required for embodied intelligence. In contrast, our benchmark achieves comprehensive coverage across all nine primary reasoning dimensions, enabling more systematic evaluation of perception, reasoning, and decision-making competencies in embodied agents. Having established the comprehensive coverage of our proposed benchmark, we next employ it to evaluate model performance and validate its effectiveness in revealing embodied reasoning disparities across different vision-language models.

\begin{figure}
    \begin{subfigure}[b]{0.5\textwidth}
        \centering
        \includegraphics[width=\linewidth]{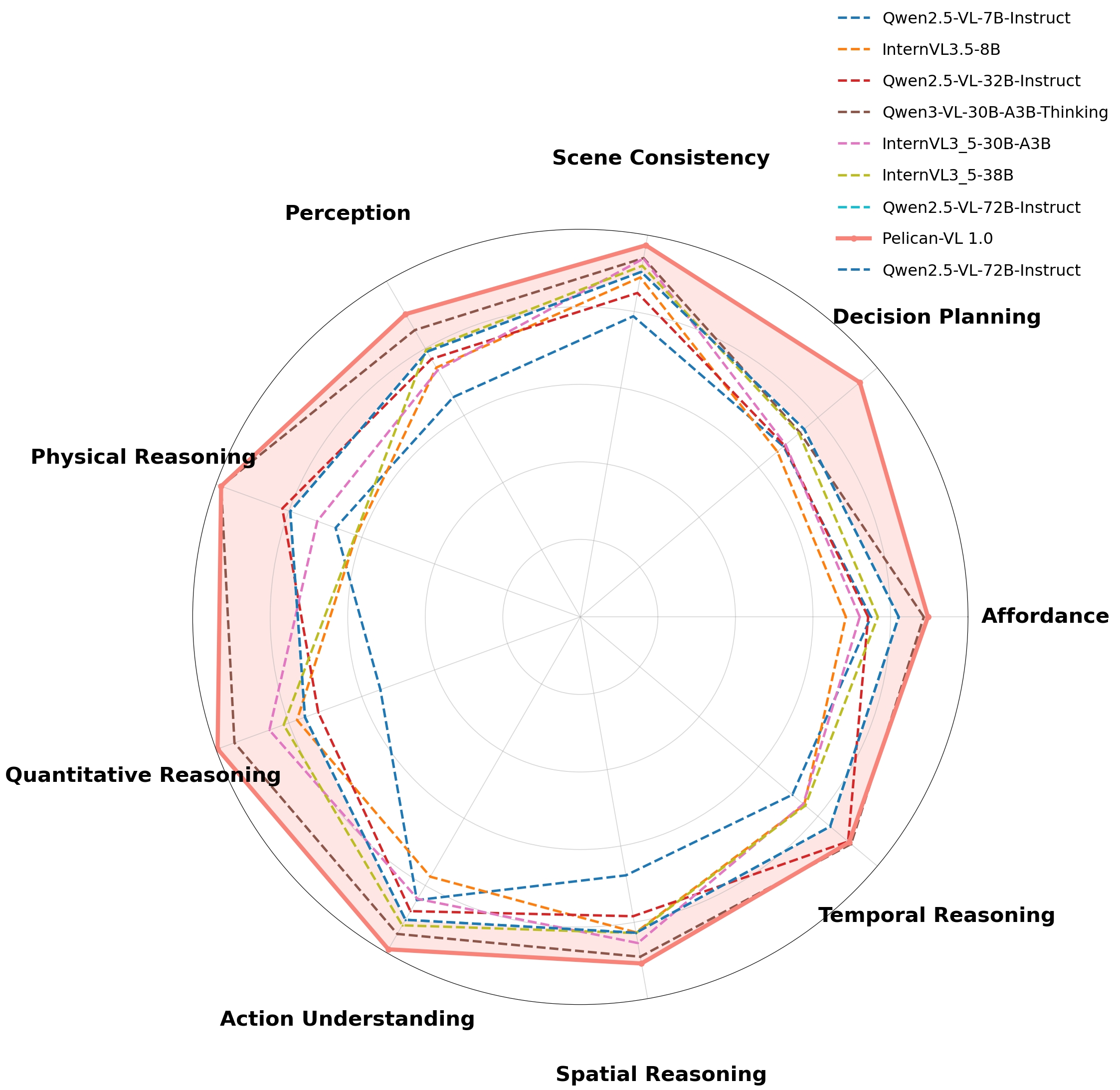}
        \caption{Pelican-VL 1.0 vs models $\leq$100B}
        \label{fig:radar_leq72b}
    \end{subfigure}
    \begin{subfigure}[b]{0.5\textwidth}
        \centering
        \includegraphics[width=\linewidth]{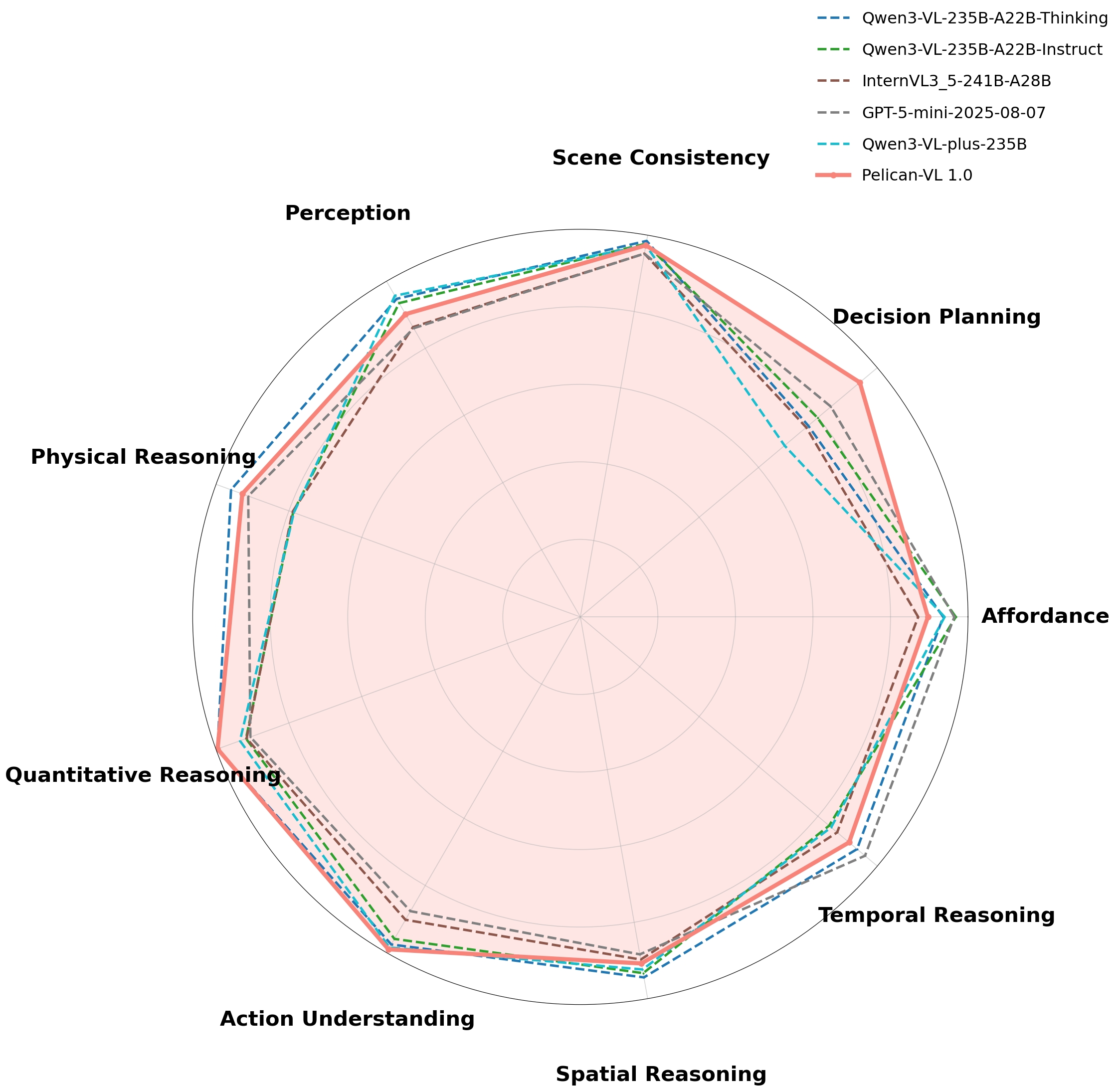}
        \caption{Pelican-VL 1.0 vs models >100B}
        \label{fig:radar_gt72b}
    \end{subfigure}

    \caption{Benchmark performance radar comparison of Pelican-VL 1.0 (72B) against other models across nine dimensions.}
    \label{fig:radar_comparison}
\end{figure}

\paragraph{Performance Comparison.} Our analysis using our new-built 9-dimension capaliblity taxonomy (Figure~\ref{fig:radar_comparison}) reveals three critical insights. First, we observe that existing open-source models exhibit uneven capability profiles. For example, models like Qwen2.5-VL-72B-Instruct show deficits in core embodied areas like Physical Reasoning and Quantitative Reasoning, even while performing well in others like Scene Consistency. This highlights that standard VLM training fails to build holistic embodied intelligence. Second, our DPPO framework is effectively solves this. Starting from a base model with similar imbalances, Pelican-VL 1.0 achieves both a comprehensively balanced capability profile and achieves SOTA performance across all nine dimensions. Third, while large-scale, proprietary models (e.g., Qwen3-VL-plus-235B) are generally more well-rounded, our Pelican-VL 1.0 not only matches their balance but also outperforms all counterparts in critical dimensions like Decision and Task Planning and Scene and Action Understanding.

Building upon this unified evaluation framework, we systematically compare Pelican-VL 1.0 with a wide range of vision-language models. As illustrated in Figure 7, Pelican-VL 1.0 demonstrates substantial and well-distributed performance gains over the Qwen2.5-VL-72B-Instruct base model, achieving an average improvement of 15.4\% across all nine reasoning dimensions. The most pronounced enhancements emerge in Quantitative and Numerical Reasoning, Physical and Causal Reasoning, and Decision and Task Planning, indicating that embodied, robot-centric fine-tuning effectively strengthens physical grounding, causal reasoning, and long-horizon planning. Steady improvements are also observed in Perception and Object Grounding, Scene Consistency, and Spatial Reasoning, suggesting enhanced perceptual-cognitive integration through multimodal alignment. When compared with models of similar or smaller scale (\(\leq 100\)B parameters)—including the Qwen2.5-VL-7B/32B/72B, Qwen3-VL-30B, and InternVL 3.5 series—Pelican-VL 1.0 maintains superior and balanced performance across all embodied dimensions. Against large-scale and closed-source systems exceeding 200B parameters (e.g., Qwen3-VL-235B, InternVL 3.5-241B, GPT-5-mini, and Qwen3-VL-plus-235B), our model remains comparable overall, outperforming all counterparts in Decision and Task Planning and Scene and Action Understanding. 

Overall, the consistent improvements across all reasoning dimensions validate both the robustness of Pelican-VL 1.0 and the effectiveness of our proposed benchmark. These findings demonstrate that an embodiment-aligned evaluation framework provides a reliable and holistic measure of embodied reasoning, bridging the continuum from low-level perception and physical understanding to high-level planning and decision-making. This unified perspective establishes a strong foundation for advancing embodied intelligence research and benchmarking future multimodal foundation models.

\subsubsection{Function Call}
In the embodied intelligence domain, models leverage their function-calling capabilities to convert inputs into executable tool-invocation operations. In practical, models need to recognize and analyze information from complex and dynamic dialogues to accurately perform function calls. 

In this section, we first fine-tuned the model using Qwen3 Tools\cite{qwen3-vl} dataset to enable the model to initially have function calling capabilities. In this way, we observe that while the model's overall performance in function calling improved significantly—with notable optimizations in both function name accuracy and parameter correctness—its general-purpose capabilities declined markedly, leaving it only able to handle function-calling tasks. To address this issue, we use GPT-4o to generate general responses for function-irrelevant information, expanding the irrelevant type data.

Furthermore, to further enhance the model's ability to understand multi-turn function calls, we built multi turn function call dataset upon the Xlam dataset\cite{liu2024apigen} that contains multiple function calls within a single-turn response. We decompose single-turn dialogues through intent recognition and boundary division, then use GPT-4o to reconstruct multi-turn dialogues based on their dependency relationships and the presence of function-related content in the answers. Finally, we train our model on this reconstructed datasets, and the experimental results are presented in Table \ref{tab:fcres}.

\begin{table*}[h]
\centering
\small
\caption{Comparison results on the Berkeley Function-Calling Leaderboard. An asterisk (*) represents that model native support for function/tool calling.}
\begin{tabular}{l|ccccccc|cc}
\toprule
\multirow{2}{*}{Model}  & Non-Live      & Live          &  Multi         & Web           & \multirow{2}{*}{Memory}    & Relevance  & Irrelevance  & Overall \\
& AST Acc       &  AST      &  Turn     &  Search   &     & Detection &  Detection & Acc \\
\midrule
GLM-4-9b-Chat* & 36.20 & 63.70 & 4.50 & 0.00 & 9.50 & 75.00 & 79.60 & 21.20 \\
Llama-3.3-70B-Instruct* & 88.20 & 76.70 & 17.90 & 9.50 & 8.20 & 100.00 & 53.80 & 30.80 \\
Gemini-2.5-Flash-Lite-Preview-06-17* & 86.40 & 67.90 & 12.50 & 27.50 & 6.50 & 43.80 & 91.30 & 35.10 \\
QwQ-32B* & 86.90 & 79.00 & 15.60 & 10.00 & 26.00 & 75.00 & 71.50 & 35.60 \\
Qwen3-30B-A3B-Instruct-2507* & 85.80 & 76.60 & 34.30 & 20.50 & 20.40 & 87.50 & 77.10 & 42.40 \\
claude-3.5-haiku-20241022* & 41.10 & 77.90 & 41.00 & 50.50 & 13.10 & 93.80 & 64.90 & 43.40 \\
GPT-4o-mini-2024-07-18* & 80.40 & 67.40 & 30.10 & 39.00 & 21.30 & 81.30 & 80.40 & 43.90 \\
Mistral-large-2411*(123B) & 85.50 & \textbf{81.50} & 21.50 & 28.00 & \textbf{43.40} & 93.80 & 68.20 & 44.30 \\
DeepSeek-V3-0324*(671B) & 88.80 & 79.90 & 33.00 & 32.50 & 22.40 & 93.80 & 74.60 & 45.20 \\
\rowcolor{pelicanmain}
Our model & \textbf{90.20} & 79.50 & 29.80 & 36.50 & 21.50 & 75.00 & 85.40 & 46.00 \\
GPT-5-nano-2025-08-07* & 66.20 & 58.90 & 26.50 & \textbf{71.50} & 25.20 & 62.50 & \textbf{89.70} & 48.80 \\
DeepSeek-R1-0528*(685B) & 75.70 & 80.90 & \textbf{44.50} & 63.00 & 0.00 & 68.80 & 73.60 & \textbf{48.90} \\
\bottomrule
\end{tabular}
\label{tab:fcres}
\end{table*}


We use Berkeley Function Calling Leaderboard (BFCL)\cite{patil2025bfcl} \footnote{A portion of the BFCL has been extracted; the complete BFCL can be seen https://gorilla.cs.berkeley.edu/leaderboard.html} to evaluates the LLM's ability to call functions (aka tools) accurately. This leaderboard consists of real-world data and will be updated periodically. As shown in the results of Table \ref{tab:fcres}, our model has achieved competitive performance compared to large-scale parameter models such as DeepSeek-R1 and GPT-5-nano. Additionally, we found that our model does not lose its general conversational capabilities even after being trained on multi-turn dialogue function call data.

\section{Downstream Applications}
This section presents a unified downstream task validation of \model across four downstream axes. Section~\ref{dst:zero} focuses on generalization—the model’s capacity to transfer and adapt to novel tasks and scenarios, establishing the foundation for flexible embodied intelligence. Section~\ref{dst:proactive} investigates the application of contact-rich tactile manipulation. It demonstrates how Pelican-VL 1.0's physical reasoning capability is leveraged within a sensorimotor loop to enable gentle and robust grasping.
Section~\ref{dst:func} highlights embodied function calling, a critical mechanism that enables robust interaction with diverse tools and modular actions within physical tasks. Finally, Section~\ref{dst:long} demonstrates the model’s capability for long-horizon reasoning and planning, revealing how sustained meta-loop training translates into advanced, context-aware embodied cognition. Each subsection builds upon and complements the others, collectively evidencing the comprehensive strengths of our approach in versatile, real-world applications.

\subsection{Zero-shot Object Manipulation with Affordance}\label{dst:zero}


\model achieves state-of-the-art performance in affordance generation across multiple benchmarks and enables zero-shot object manipulation without fine-tuning or human demonstrations. To validate its real-world applicability, we deploy it on a tabletop single-arm robotic platform.
The model is hosted on a cloud service that receives sensory data from the robot and returns structured JSON responses to the local controller. The local controller acquires RGB image observations and executes the generated joint trajectories on a parallel-jaw gripper arm.

\begin{figure}
	\centering
	\includegraphics[width=\linewidth]{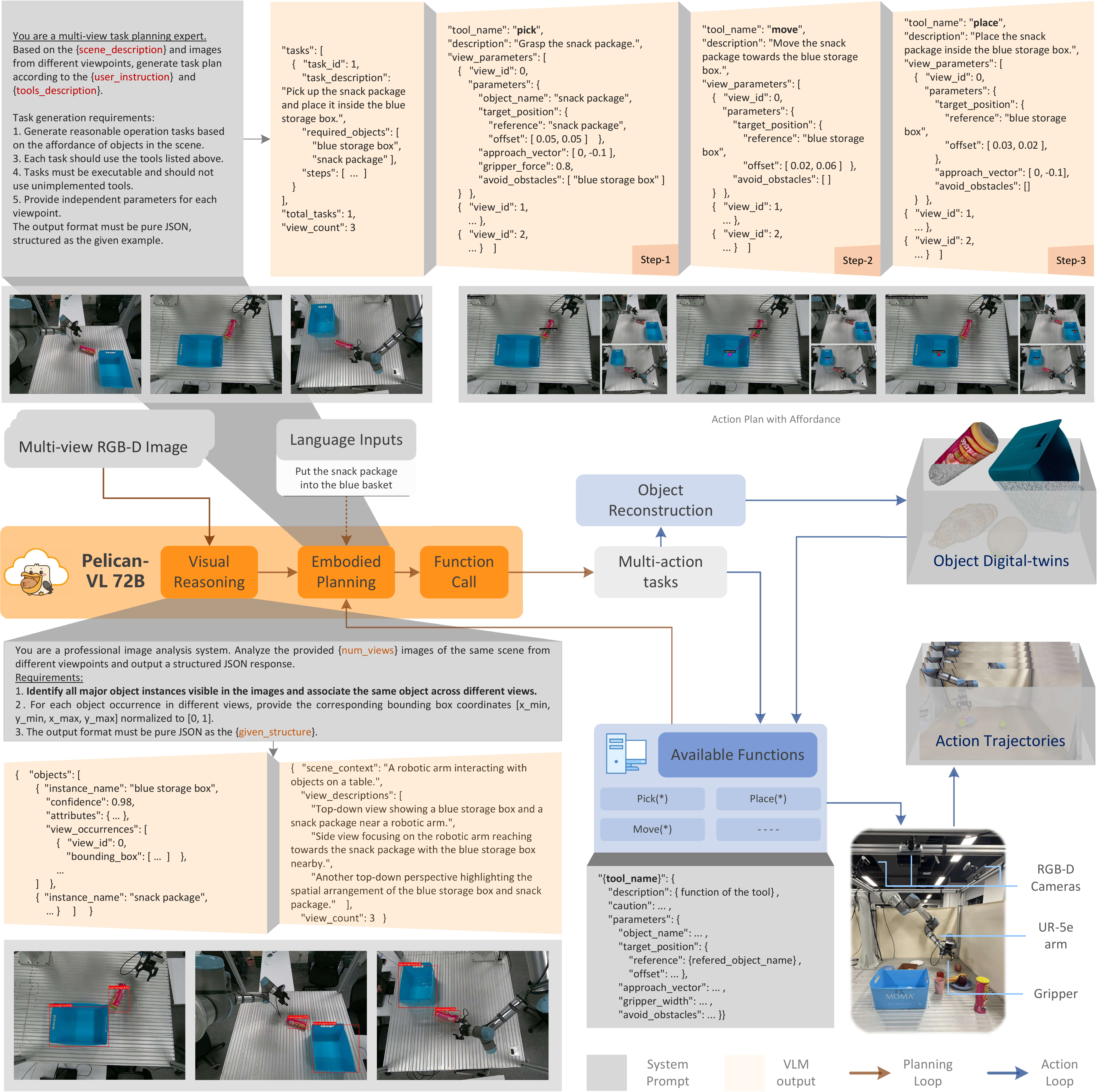}
	\caption{\model as affordance-based robot action planner with multi-view input. \model first generates visual reasoning for the inputs, then a structured action plan with affordance is generated according to optional language inputs and available functions of the robot. By using manipulation tool with generalization ability, the system achieves zero-shot object manipulation. In function calling phase, reconstructed object digital-twins and action trajectories are saved.}
    \label{fig:affordance:module_structure}
\end{figure}

As shown in \autoref{fig:affordance:module_structure}, the cloud backend employs the 72B-parameter \model for behavior-level task planning and coordination through affordance-driven reasoning. The system follows a hierarchical control structure:
\begin{enumerate}
\item \textbf{Visual Perception:} Multi-view RGB image inputs are first processed to extract structured representations for each object, including attributes, affordances, and spatial locations.
\item \textbf{Embodied Planning:} The extracted information is fed back to the VLM for a second-stage VQA process, where high-level manipulation objectives are decomposed into detailed action sequences.
\item \textbf{Function Calling:} Using its function-calling capability, the VLM translates planned actions into executable manipulation tool calls. These calls are executed locally, where a motion planner generates corresponding joint trajectories and gripper actions.
\end{enumerate}

\subsubsection{Zero-Shot Manipulation Strategy based on Affordance Reasoning and Multi-View Consistency\label{subsubsection:zero-shot-manipulate}}
Approaches for zero-shot robot manipulation with VLMs can be categorized into: (1) direct code generation, (2) affordance-based function calling, and (3) direct trajectory generation via vision-language-action (VLA) models. We adopt the affordance-based function-calling approach due to its dynamic tool reloading capability, which preserves generalization while allowing incremental skill expansion.
Direct code generation requires expert verification before deployment, while VLA models demand additional trajectory data and fine-tuning to generalize across new tasks or embodiments. Moreover, tasks requiring high precision or compliant motion remain challenging for current VLA frameworks.

\begin{figure}
	\centering
	\includegraphics[width=0.9\linewidth]{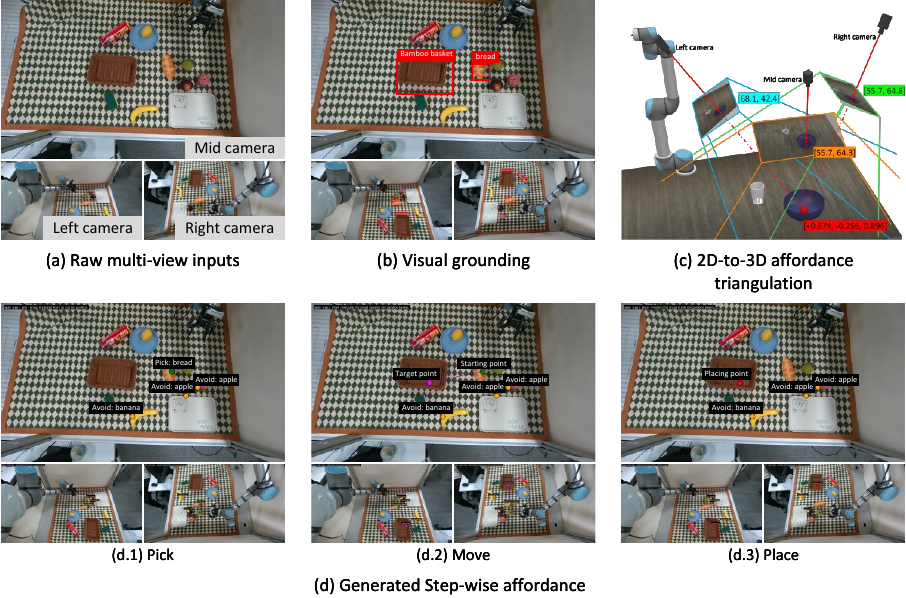}
	\caption{\model can generate consistent affordances for multi-view inputs, so that 2D affordances can be triangulated into 3D target points as direct robot action commands. (a) For triangulation, as least three views are needed for inputs. (b) Visual groundings are first generated for objects being manipulated. (c) 2D pixel coordinates and camera extrinsics are needed for triangulation. (d) For a pick-place task, we generate pick target point, moving direction, place target point, and avoiding objects for each step.}
    \label{fig:affordance:multi_view_consistency}
\end{figure}

During task planning, \model produces intermediate visual-grounding and affordance results in structured text, effectively decoupling visual reasoning, embodied planning, and function execution. This design improves stability over single-step function generation from raw visual inputs, analogous to chain-of-thought prompting for enhanced reasoning transparency.
To address spatial ambiguity inherent in single-view affordance estimation, \model is trained on densely annotated multi-view video frames, ensuring consistency in affordance generation and spatial reasoning. By incorporating three or more calibrated RGB views, the system achieves geometrically consistent 3D task definitions. Given camera extrinsics, accurate 3D manipulation targets are reconstructed by intersecting corresponding 2D camera rays, as visualized in \autoref{fig:affordance:multi_view_consistency}.

\subsubsection{Autonomous Data Collection and Learning}

The system supports both user-instructed and autonomous modes. With a language input, \model decomposes the given instruction into sequential manipulation actions. Without user input, it autonomously generates random tasks based on inferred affordance results, enabling fully autonomous operation for large-scale data collection.
During execution, both object-centric (e.g., reconstructed meshes, point clouds) and robot-centric (e.g., joint positions, end-effector poses, gripper states, and force readings) data are recorded. These multimodal datasets are employed for RL and imitation learning (IL) in downstream manipulation tasks.

In RL applications, diffusion-based models are adopted for mesh and material generation during scene reconstruction. Conditioned on the affordance and attribute descriptions from \model, the generated meshes and textures exhibit high realism. Additional material priors can be incorporated in simulation to improve real-to-sim fidelity and enhance sim-to-real transfer. The collected trajectories also serve as expert demonstrations for initializing replay buffers, thereby accelerating RL convergence.

For IL, the data collection pipeline ensures perceptual alignment between the large model and recorded data. Unlike traditional teleoperation-based datasets, which suffer from perception-action misalignment between human operators and model reasoning, our framework delegates visual reasoning to \model and trajectory generation to the motion planner. This consistent division of labor ensures alignment between training and deployment, resulting in superior generalization across domains with varying lighting and clutter conditions.

\subsubsection{Generalization Abilities}

\begin{figure}
	\centering
	\includegraphics[width=0.9\linewidth]{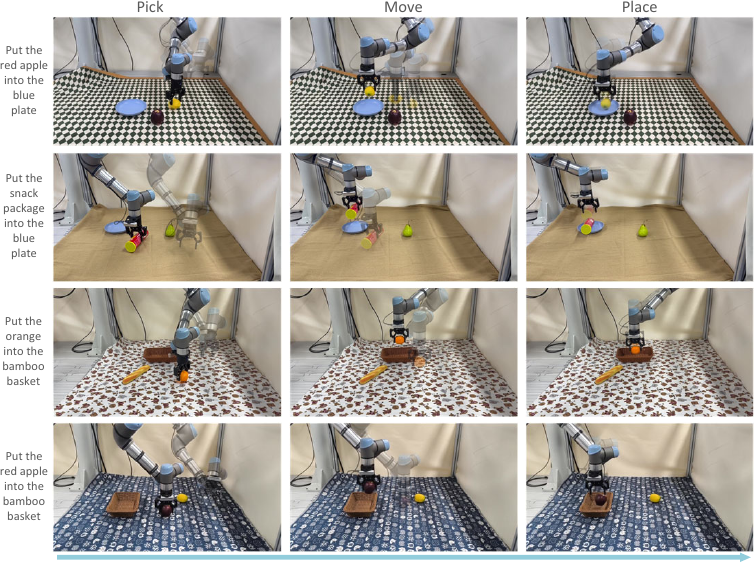}
	\caption{Generalization to tasks with different objects and scenes.}
    \label{fig:affordance:scene_generation}
\end{figure}

As demonstrated in \autoref{fig:affordance:multi_view_consistency}, \model generates accurate multi-view visual grounding and affordance-based task decompositions. Grasp points are highlighted in green, avoidance regions in yellow, and placement targets in pink. The decompositions are consistent across viewpoints, and the integrated 2D results yield 3D action targets, as shown in \autoref{fig:affordance:multi_view_consistency}. The affordance-based manipulation framework generalizes effectively to novel objects and novel scenes, as illustrated \autoref{fig:affordance:scene_generation}.

\subsection{Closing the Sensorimotor Loop: Proactive Prediction and Reaction Adaptation}\label{dst:proactive}

Human-like grasping requires both proactive prediction and reactive adaptation. As illustrated by Johansson and Flanagan \cite{johansson2009coding}, the process operates on a continuous cycle, prediction - tactile adaption - memory update. (1) Before contact, the brain uses visual cues to identify the object and recall sensorimotor memories, forming a prior estimate of its physical properties. This allows for the pre-parameterization of motor commands, enabling a appropriately scaled grip force from the moment of fingertip contact. (2) During manipulation, fast-conducting tactile afferents in the fingertips provide real-time feedback on micro-slips and force distribution. This signal is continuously updated for corrective adjustments to the grip force, ensuring stability while avoiding excessive grip force. (3) Following the interaction, the discrepancy between the predicted and actual sensory outcome is used to refine the internal model of the object. This updated memory is then retrieved to improve the accuracy of the predictive command in all future interactions, creating a cycle of continuous learning.

\begin{figure}
    \centering
    \includegraphics[width=\linewidth]{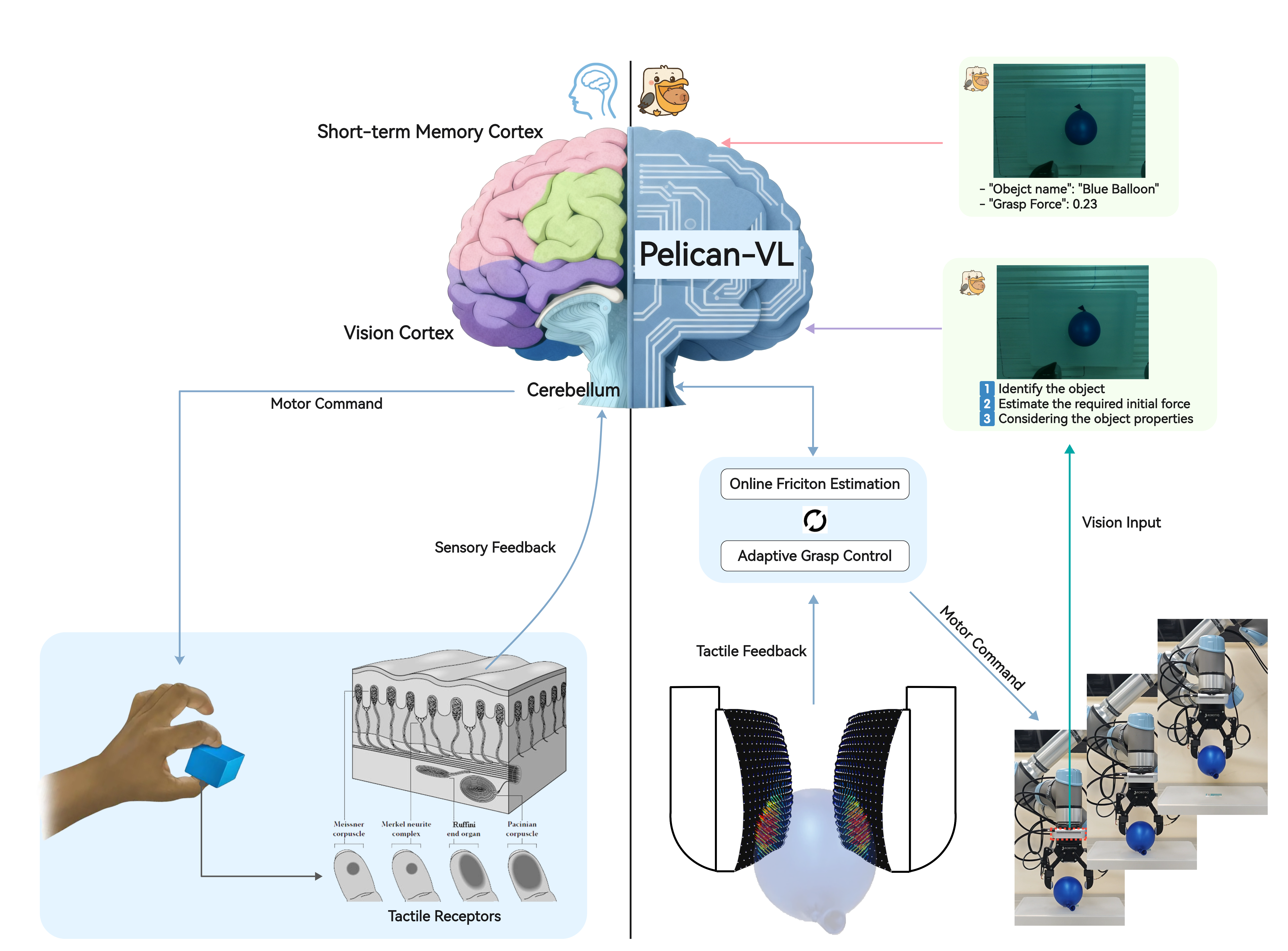}
    \caption{An integrated architecture for adaptive robotic grasping. The pipeline begins with the \model predicting an initial grasp force from visual input. This prior activates a synchronized estimation-control module, where a particle filter continuously refines the friction estimate, and a reactive controller adjusts the grip force accordingly. The system concludes by storing successful interactions in a memory graph, creating a closed-loop of proactive prediction, haptic adaptation, and experiential learning.}
    \label{fig:VlmGraspFramework}
\end{figure}

As shown in Figure \ref{fig:VlmGraspFramework}, we propose a novel framework in robotic grasping that replicates above three-component architecture leveraging the real-world physical reasoning capability of \model, synchronized online friction estimation and adaptive grasp control, and a physical-memory system.


\begin{figure}
    \centering
    \includegraphics[width=0.8\linewidth]{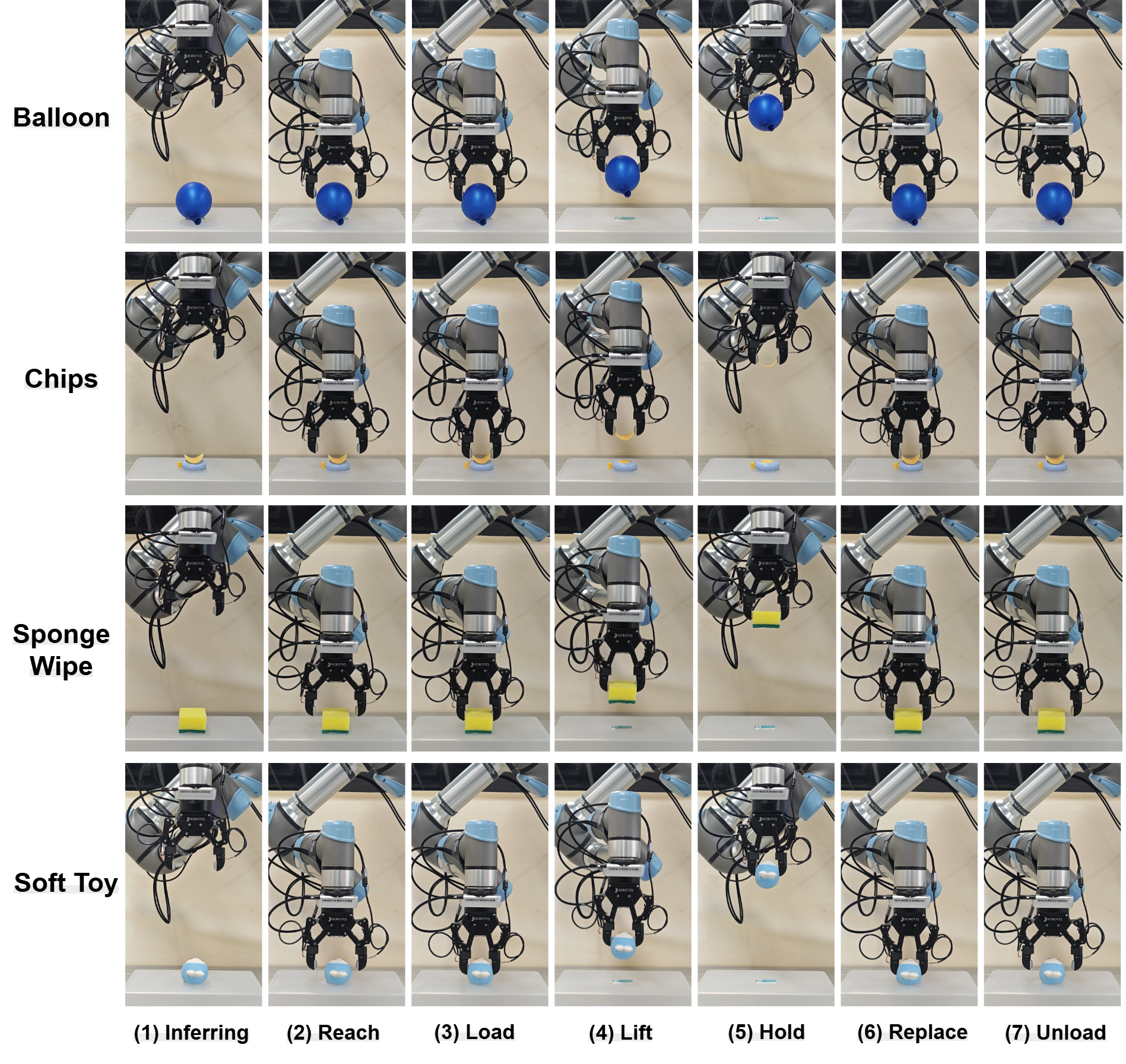}
    \caption{Example tasks demonstrating our sensorimotor framework grasping delicate and compliant objects. The initial force prior provided by \model accelerates the convergence of the online adaptive controller. This creates a continuous learning cycle, where the results of each physical interaction are stored in a knowledge graph, refining the system's priors for future interactions and enhancing long-term performance.}
    \label{fig:grasp results}
\end{figure}

\subsubsection{Robust Gentle Grasp for Delicate and Compliant Objects}

Figure \ref{fig:grasp results} presents a real-world robotic evaluation of the proposed sensorimotor framework for grasping delicate and compliant objects. We validate the complete, seven-phase grasping process as outlined in our methodology.

\begin{itemize}
    \item \textbf{Reach Phase}: The gripper closes until the tactile sensors register a normal force matching the prior, $F_{n}^{init}$, predicted by the \model.
    \item \textbf{Load Phase}: The synchronized process of online friction estimation and adaptive grasp control is initiated, refining the grip force based on real-time haptic feedback.
    \item \textbf{Lift Phase}: The robot elevates the object to a predefined target height. Throughout this motion, the grasp force is continuously adapted to maintain stability.
    \item \textbf{Hold Phase}: The object is held steadily at the target position, demonstrating the controller's ability to maintain a stable grip.
    \item \textbf{Transport \& Return} (Phases 6-7): The object is smoothly transported and returned to its original position, completing the manipulation cycle.
\end{itemize}

This sequence clearly demonstrates the framework's capability for dynamic force adjustment and stable manipulation of challenging objects.

\subsection{Embodied Function Call}\label{dst:func}

\begin{figure}[htbp]
    \centering
    \includegraphics[width=\linewidth]{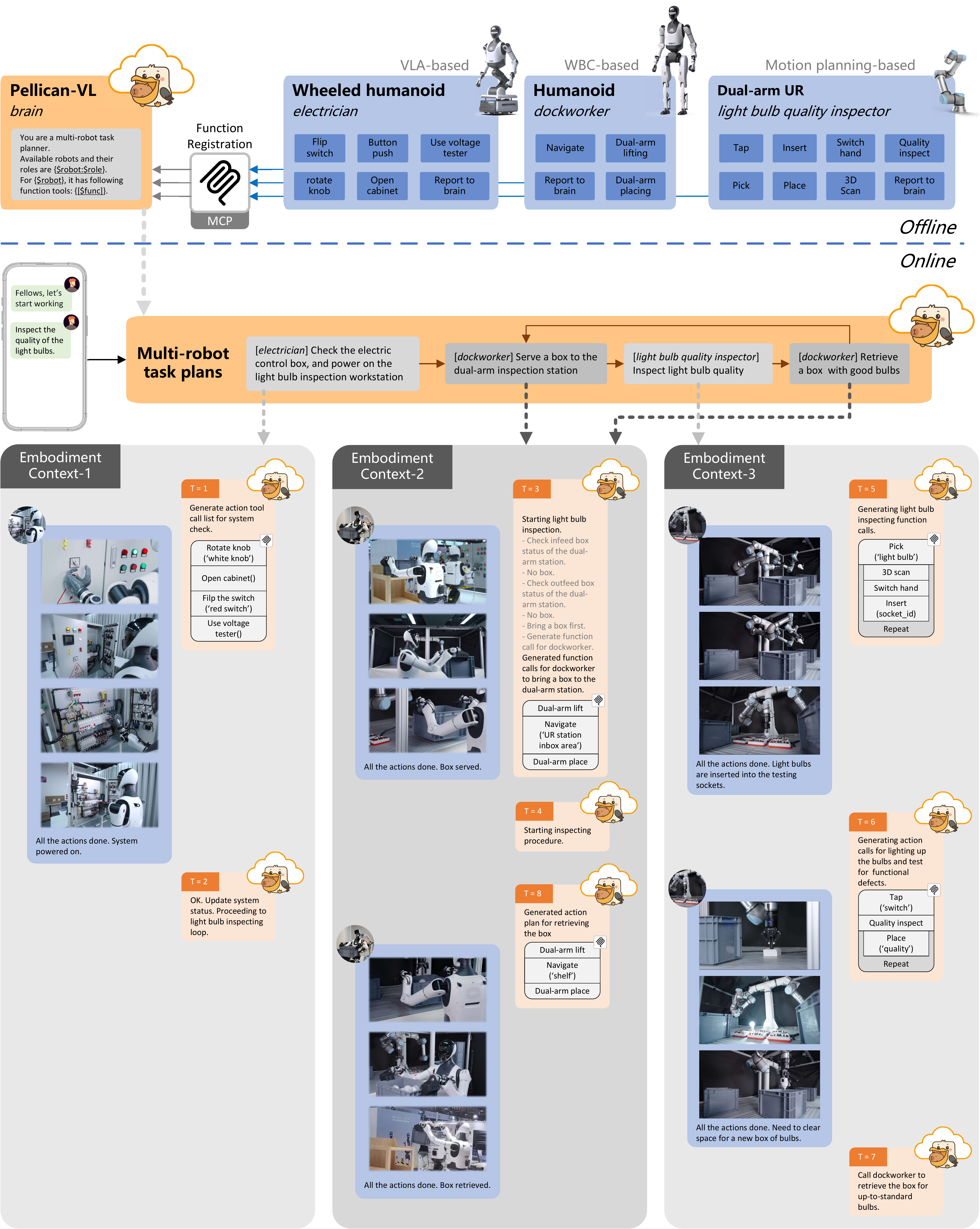}
    \caption{\model for multi-robot collaboration task via robust function calls in multi-turn conversation. Available functions are cached during the offline phase, through MCP (Model Context Protocol). Task plans are detailed first as behavior plans then action function calls in multi-round conversation (T for conversation tag) for different embodiment context.}
    \label{fig:function_call:function_call_demo_structure}
\end{figure}

In real-world robotic applications, system-level intelligence often requires both inter-robot collaboration and human–robot interaction capabilities. To evaluate and demonstrate our model’s performance under such conditions, we developed a multi-robot system (as shown in \autoref{fig:function_call:function_call_demo_structure}) built upon \model, illustrating how function calls operate within cross-robot collaborative pipelines.

The system encompasses three distinct robotic embodiments (a wheeled humanoid robot, a bipedal humanoid robot, and an industrial manipulator) and three types of control tool implementations (VLA, WBC (whole-body control), and traditional motion planning control). In the offline stage, \model fetches the system prompts, character profiles, and functional tools of each embodiment via the MCP (Model Context Protocol) to construct the system context. Caching all functional tools enables faster response during real-time execution. In the online stage, \model first receives task instructions from the user interface and decomposes the given system-level task into multi-robot task plans, where each step corresponds to a behavior-level task for a specific embodiment. During asynchronous execution, each embodiment’s behavior-level task is further decomposed by \model into parameterized action function calls.

The function call–based embodied system built on \model was further employed to rapidly construct an industrial light bulb inspection pipeline. As shown in \autoref{fig:function_call:function_call_demo_structure}, after the user issues a command to start the process, the behavior-level tasks are decomposed as follows: powering on the system, which serves as a prerequisite task handled by the electrician; and a cyclic process involving light bulb inspection and sorting, as well as box handling, performed by the inspector and the dockworker. The corresponding function calls are generated via the MCP protocol in a multi-round dialogue manner across different embodied entities, as follows:
\begin{enumerate}
\item \textbf{Check system and power on.} [T=1] \model generates function calls for the wheeled humanoid to check the electrical control cabinet and power on the inspection station. VLA functions for the dexterous hand are called with language inputs generated by \model.
\item \textbf{Load and unload the inspection station.} [T=3, T=8] Before starting the light bulb inspection procedure, a CoT reasoning process determines the next system behavior — whether to load the inspection station with uninspected bulbs, unload it with inspected bulbs, or start inspection with an already loaded box of uninspected bulbs. If loading or unloading is required, whole-body control (WBC) functions for dual-arm manipulation and navigation are called with parameters generated by \model.
\item \textbf{Quality inspection.} [T=5, T=6] Upon receiving a box of uninspected bulbs, \model generates function calls for the dual-arm station to perform structural and functional inspections on the incoming bulbs. Inspected bulbs are sorted into two boxes — one for qualified products and the other for defective ones — which are then unloaded by the dockworker robot.
\end{enumerate}

With the help of a unified interface protocol, our model can be further integrated with a wider range of embodied and general-purpose tools (such as map navigation and data management), enabling the rapid development of more diverse application scenarios and the realization of more autonomous cyber-physical intelligent agents.

\subsection{Long-Horizon Task Reasoning and Planing}\label{dst:long}

\begin{figure}[h]
    \centering
    \includegraphics[width=0.9\linewidth]{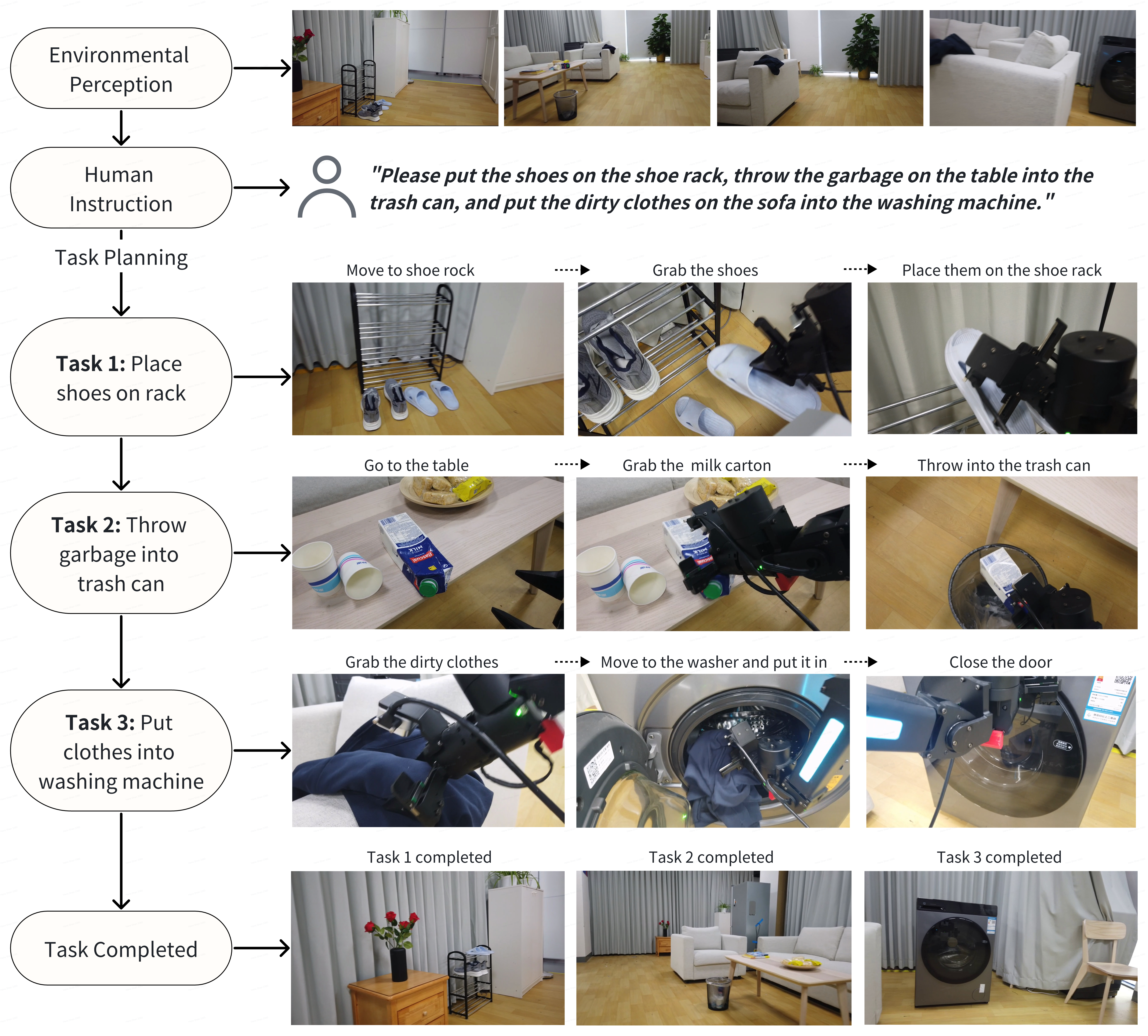}
    \caption{Demonstration of \model executing a long-horizon embodied task in a real domestic environment. The model interprets a natural-language instruction and autonomously performs sequential actions—placing shoes on the rack, disposing garbage into the trash can, and loading clothes into the washing machine—through integrated perception, spatial reasoning, and adaptive planning across multiple regions.}
    \label{fig:54}
\end{figure}

To further evaluate \model’s capability in long-horizon planning and embodied reasoning, we conducted a comprehensive experiment in a real domestic environment, as shown in Figure \ref{fig:54}. The task required the model to interpret a natural-language instruction and autonomously complete a sequence of manipulation and navigation actions without explicit step-by-step guidance.

The environment consisted of a typical living-room scene containing a shoe rack, a sofa, a coffee table, and a washing machine, along with everyday objects such as shoes, empty cups, milk cartons, and clothes. The human command was:\textbf{\textit{Please put the shoes on the shoe rack, throw the garbage on the table into the trash can, and put the dirty clothes on the sofa into the washing machine.}}  This instruction involves multi-stage perception, spatial reasoning, and dynamic task sequencing across several spatially separated regions.

During execution, \model first constructed a semantic representation of the scene through its perception module, identifying key objects and their spatial relations. It then autonomously planned the action sequence, moving successively to the shoe rack, table, and washing-machine areas, and executed multi-modal interactions such as grasping, carrying andplacing. Throughout the task, the model maintained situational awareness, continuously updated its internal state representation, and adapted its plan based on real-time sensory feedback.

Ultimately, the robot successfully placed the shoes on the rack, disposed the garbage from the table into the trash can, and transferred the clothes into the washing machine, closing the lid upon completion. These results highlight \model’s ability to comprehend and execute complex, compound human instructions, demonstrating coherent understanding of spatial relations and long-horizon task control in the physical world. This experiment provides strong empirical evidence of the model’s embodied reasoning competence and real-world robustness.

\section{Related Work} 
\subsection{Embodied Foundation Model}
The central challenge in embodied AI is the creation of a generalist Embodied Foundation Model—a scalable robotic brain that unifies perception, reasoning, and action. By its very definition, such a brain is innately multimodal. This pursuit has therefore been directly catalyzed by the profound advancements in large VLMs~\cite{zhang2024vision,liu2025nexus,team2023gemini,wen2025infmasking}, which provide the foundational architecture for this synthesis. As noted in our introduction, the field has largely bifurcated into two primary strategies to achieve this. The first strategy is heterogeneous data scaling, where major labs demonstrate high performance by adapting massive, pre-trained VLMs through data pyramid scaling. This data-scaling paradigm is exemplified by flagship efforts such as Google's Gemini Robotics~\cite{abdolmaleki2025gemini}, which leverages its powerful VLM for complex reasoning in physical tasks; NVIDIA's GR00T N1~\cite{bjorck2025gr00t} \& Cosmos-Reason1~\cite{azzolini2025cosmos}, open models explicitly designed for generalist humanoid control; ByteDance's GR-3~\cite{cheang2025gr}, which integrates web-scale knowledge; Physical Intelligence's $\pi_{0.5}$~\cite{intelligence2504pi0}, which uses novel flow-based architectures for open-world generalization; RoboBrain~\cite{Ji2025RoboBrain} and others~\cite{figure2025helix,fan2025xr}. While these approaches have pushed the boundaries of performance, they often rely on the sheer scale of the data, lacking a sophisticated algorithmic framework to manage, refine, and consolidate learning from such vast, heterogeneous, and often noisy sources. The second strategy is architectural refinement, which proposes more decoupled, layered systems. This approach, seen in models like Figure's Helix and Wall-OSS~\cite{zhai2025igniting}, typically uses a large VLM as a high-level ``brain" for reasoning and task planning, which in turn commands a separate, smaller, reactive policy for low-level motor control. While architecturally elegant, this paradigm often operates on smaller, more specialized datasets, thus lacking the sheer scale and data diversity required for true, open-world generalization. 

This bifurcation clearly highlights the industry's consensus: Vision-Language-Action (VLA) models (for manipulation) and Vision-Language-Navigation (VLN) models (for mobility) are increasingly treated as downstream applications. These specialized models typically take a powerful, pre-trained VLM as a foundational backbone and are then fine-tuned for their specific domains—VLA focusing on contact-rich, fine-grained control, and VLN on long-horizon spatial reasoning. Our work addresses the critical missing piece identified by this bifurcation: the intelligent, adaptive learning mechanism that unifies massive scale with a robust architecture. We are focused on creating the general-purpose Embodied Foundation Model itself. 

\subsection{Post Training Algorithm} 
Post-training algorithms are essential for aligning foundational models. The baseline approach, SFT~\cite{tie2025survey,gui2024survey}, is a stable mechanism for static knowledge infusion from expert data. While effective for domain adaptation, it is inherently limited by its pre-collected dataset and cannot discover novel policies. This limitation led to the integration of RL~\cite{wu2025reinforced,zhai2024rl4vlm} algorithms. The field evolved from complex, online reward-modeling methods like PPO~\cite{schulman2017proximal} to more stable, offline alignment techniques. DPO~\cite{rafailov2023direct} marked a significant algorithmic shift by reframing alignment as a direct preference-learning problem. This was further refined by algorithms like GRPO~\cite{shao2024deepseekmath}, which generalizes DPO to leverage richer, ranked preference signals instead of just binary pairs.

Despite this algorithmic progress, a fundamental gap remains: SFT (for knowledge) and RL/DPO/GRPO (for alignment/discovery) are still treated as separate, disjointed, and often one-off phases. This fragmentation is a critical bottleneck for embodied intelligence, which demands a continuous cycle of both skill discovery and stable knowledge consolidation.

Our work directly addresses this fragmentation. Orchestrated by our Metaloop mechanism, DPPO is not just another alignment algorithm; it is an algorithmic scheduler. It dynamically alternates between an Exploratory Grounding loop (leveraging RL/GRPO) to actively discover and refine novel skills, and a Targeted Remediation loop (leveraging SFT) to rapidly and stably consolidate these discoveries. This synthesis unifies the exploratory power of RL with the stable consolidation of SFT, enabling robust, continuous self-improvement.

\subsection{Embodied Training Data}
Training datasets for embodied large models have undergone a profound transformation from 2020 to 2025, evolving from small-scale simulation benchmarks to expansive multimodal repositories, with paradigms shifting toward hybrid sources to bridge sim-to-real gaps and enhance cross-embodiment transfer. Early simulation datasets, such as ACRONYM~\cite{eppner2021acronym} with 17.7M grasp poses and ManiSkill2~\cite{gu2023maniskill2} offering 4M visual frames for dexterous manipulation, provided scalable yet physics-limited environments for policy pretraining, enabling rapid iteration but exposing limitations in real-world noise robustness (e.g., >50\% sim-to-real failure rates). Real-world trajectory collections, exemplified by RoboNet~\cite{dasari2019robonet}'s 15M RGB frames and BridgeData V2~\cite{walke2023bridgedata}'s 60k multimodal episodes, incorporating audio and haptics, captured authentic dynamics for imitation learning but suffered from high collection costs and modality sparsity, restricting scalability to mid-sized corpora. By 2024-2025, multimodal and hybrid datasets like Open X-Embodiment~\cite{vuong2023open} with over 1M trajectories across 22 robot embodiments and 527 skills—integrating vision, language, and proprioception—democratized diverse, collaborative resources, while recent advances such as AgiBot World~\cite{bu2025agibot}, RoboMIND~\cite{wu2024robomind}, and EO-1.5M~\cite{qu2025embodiedonevision} fused 1.5M interleaved vision-text-action samples to support unified decoders, achieving 10-100x scale gains over prior works. Recent advancements in embodied training datasets from 2024 to 2025 have further enriched multimodal repositories, for example, RoboScape~\cite{shang2025roboscape} provides high-resolution RGB-depth sequences in physics-informed environments, Embodied-Spatial-84K~\cite{yuan2025embodied} dataset mitigate limitations in prior works' scene variability and enabling robust cross-embodiment transfer.

\subsection{Embodied Benchmark}
Benchmarks for embodied large models rapidly evolved from siloed, high-level simulations like Habitat~\cite{ku2020room} and ALFRED~\cite{shridhar2020alfred} to low-level dexterity tests like ManiSkill2~\cite{gu2023maniskill2} and LIBERO~\cite{liu2023libero}. These early platforms were critical for prototyping but highlighted significant sim-to-real gaps and low-noise transfer failures, proving that simple task success was an insufficient metric.

This spurred a move toward comprehensive, multitask frameworks. Open X-Embodiment~\cite{vuong2023open} enabled cross-embodiment leaderboards, culminating in 2025 with holistic platforms like Embodied Arena~\cite{ni2025embodied} and its extensions like Isaac Lab-Arena. These massive suites demonstrated SOTA performance (e.g., 90\%+ long-horizon via Gemini Robotics 1.5~\cite{team2023gemini}), proving that models can succeed at complex, long-horizon tasks.

However, in this work, we conducted a deep analysis of these key benchmarks and identified a fundamental flaw. While holistic platforms like Embodied Arena show if a model succeeded, they fail to show why it failed. This gap has led to a recent, fragmented explosion of diagnostic benchmarks, such as EmbSpatial-Bench~\cite{du2024embspatial}, RoboSpatial~\cite{song2025robospatial}, and OmniSpatial~\cite{jia2025omnispatial} for spatial-physical reasoning, and COSMOS~\cite{azzolini2025cosmos} or ERQA~\cite{abdolmaleki2025gemini} for CoT-reasoning. While EmbodiedBench~\cite{yang2025embodiedbench} introduced this idea of capability dimensions, the field now consists of either coarse, Pass/Fail tests (like Arena) or fragmented, niche capability tests (like RoboSpatial). There is no benchmark that diagnostically evaluates these core capability axes in an integrated way. To address this critical gap, our follow-up work will release a new benchmark designed to do exactly that.

\section{Discussion and Future Work}
\label{sec:discussion}
In this report, we introduce \model, to our knowledge, the largest open-source embodied brain model to date, along with DPPO, the novel metacognitive framework used to train it. Our results demonstrate the effectiveness of this heavy-lifting approach, achieving SOTA performance validated through extensive real-world hardware experiments. These include (1) contact-rich tactile manipulation, where Pelican is the first VLM to close the sensorimotor loop; (2) complex task-oriented affordance reasoning; and (3) industry-first long-horizon planning as a unified one brain for multi-agent systems.

More importantly, the DPPO framework validates a powerful and scalable path for iterative model improvement. By open-sourcing our complete models and toolchain, we provide a foundation that the community can easily reproduce to foster collaborative development. However, as we discuss, \model and DPPO are not an end, but rather the foundation for a much larger, long-term vision of autonomous, self-evolving intelligence. Our future work, proceeding along four critical axes, outlines a complete roadmap toward this goal.

\paragraph{Beyond Coarse Tasks: The Need for Diagnostic and Fair Platforms.}
A core discovery in our experiments is that current benchmarks are a fundamental bottleneck to self-evolve, plagued by two distinct problems. First, they lack diagnostic granularity. Benchmarks organized by \textit{dataset} or \textit{task} (e.g., ALFRED, ManipBench) provide coarse, Pass/Fail signals. They can show \textit{if} a model failed, but not \textit{why}. This black-box evaluation makes it impossible to guide targeted, efficient iteration. Second, our results highlight that embodied evaluation lacks a fair, reproducible, and trusted platform. This fragmentation forces researchers to expend significant effort simply to replicate evaluation setups, hindering fair community-wide comparison. Therefore, our future work will proceed in two directions: (1) creating a new generation of diagnostic benchmarks organized around fine-grained \textit{capability dimensions} (e.g., spatial reasoning, causal inference). This is the only way to provide the granular feedback a self-detect loop needs. (2) We will also organize and release our complete evaluation suite, aiming to provide a one-click platform for standardized, fair, and reproducible results to accelerate community progress.

\paragraph{From Petabytes of Data to Aligned Knowledge.}
This new benchmark is the prerequisite for our third goal: transforming data collection. While we aggregated massive, petabyte-scale data, it is, like others (e.g., RoboMIND, Open X-Embodiment), organized by its \textit{source}. This is inefficient for targeted enhancement. Our future work will involve building a \textit{capability-aligned} training corpus, where data is tagged by the specific model weaknesses it is designed to remediate. This creates the essential link: the diagnostic benchmark will identify a capability gap, and this new data corpus will provide the precise medicine for the \texttt{self-refine} loop.

\paragraph{Closing the Embodied Last Mile.}
We recognize that a brain (the EFM), no matter how well-trained on static datasets, is insufficient for true embodied intelligence. The last mile of self-evolution is closing the loop with physical \textit{embodiment}. Our future work must integrate the DPPO brain with both high-fidelity simulation and real-world robotic hardware. This closed loop, where the embodiment can actively interact with the environment, generate novel data, and trigger the DPPO \texttt{metaloop} for on-the-fly updates, represents the ultimate architecture for continuous improvement.

\paragraph{Igniting the ChatGPT Moment for Robotics.}
Finally, the true measure of embodied intelligence is its ability to solve critical, real-world problems. Our ultimate goal is to validate this entire \texttt{self-evolve} ecosystem---the autonomous DPPO, the diagnostic benchmark, the capability-aligned data, and the closed hardware loop---in key societal scenarios. We will target fundamental challenges in manufacturing, logistics, or healthcare. We hypothesize that the successful demonstration of such a fully autonomous, self-Evolving agent that solves a meaningful problem will represent the true ``ChatGPT moment" for robotics, finally unlocking its potential to revolutionize productivity.


\pagebreak
\bibliography{biblio}
\bibliographystyle{abbrv}
\pagebreak
\section{Contributions}\label{contributions}
Our contributors are organized based on their roles and magnitude of contribution. The specific roles are as follows:
\subsection{Core Contributors}
\begin{itemize}
    \item Yi Zhang, Che Liu, and Xiancong Ren
\end{itemize}
\subsection{Contributors}
\begin{itemize}
    \item Hanchu Ni, Shuai Zhang, Zeyuan Ding, Yingji Zhang, Jiayu Hu, Haozhe Shan, Zhenwei Niu, Zhaoyang Liu, Shuang Liu, Yue Zhao, Junbo Qi, Yan Bai, Qinfan Zhang, Dengjie Li,  Yidong Wang, Jiachen Luo, Zenglin Xu, Bin Shen, and Qifan Wang
    \item Baidu AI Cloud Baige Team (aihc@baidu.com)
\end{itemize}
\subsection{Tech Lead}
\begin{itemize}
    \item Yong Dai
\end{itemize}
\subsection{Corresponding Authors}
\begin{itemize}
    \item Jian Tang and Xiaozhu Ju
\end{itemize}

\subsection{Acknowledgement}
We would like to express our sincere gratitude to all our colleagues at X-Humanoid for their invaluable support and insightful discussions throughout this ambitious project. 
We are particularly grateful to Wanxin Tian from the Agent team; his persistent feedback and deep insights into the agent's iterative development were instrumental in shaping our research. 
Our heartfelt thanks also go to Fangzhe Yan, Wenwen Han, Zexin Yang, Ouyang Li and the entire data annotation team. Their diligent and meticulous work in processing and labeling our petabyte-scale dataset was foundational to the success of Pelican-VL 1.0. 
\section{Appendix}\label{testset}

\subsection{Grasp Contact Model and Adaptive Grasp}
\label{appendix:grasp model}
Human can sense the friction information on contact surface based on the slip state of the fingertips to control the grasping force \cite{johansson1987signals}. Therefore, slip detection is important for grasp performance. A promising approach for slip detection is to quantify the incipient slip which means to detect the distance from the macro slip (where the contact surfaces of two objects have completely slid with respect to each other) \cite{sui2023novel}. Sui \cite{sui2023novel} and Li \cite{li2025learning} proposed a contact factor based on the vision-based tactile sensor (VBTS) to quantify incipient slip and enable reactive grasp control. In this framework, we proposed a stochastic model based on the contact factor and realized a synchronized online friction estimation and adaptive grasp control.

The contact coefficient is defined as:
\begin{equation}
\label{eq-contact coefficient}
    cf = 1 - \frac{F_{\text{MER},t}}{\mu \cdot F_{\text{MER},n}}
\end{equation}
where $F_{\text{MER},t}$ and $F_{\text{MER},n}$ represent the integral of tangential and normal force magnitudes over the contact surface, respectively. The contact coefficient, denoted as $cf$, is bounded between 0 and 1. A value of $cf$ close to 0 indicates that the contact state is to macro slip, while values closer to 1 reflect greater contact stability. This metric provides a quantitative measure of grasp security and can be directly incorporated into adaptive grasping control.

In the above contact model, the friction coefficient $\mu$ is essential to be given as a prior. However, this prerequisite is often infeasible to get for arbitrary objects. This limitation has led to solutions misaligned with the goal of human-like adaptability. For instance, Li \textit{et al}. \cite{li2025learning} employ a two-stage estimation and control process that is prohibitively slow. While the online estimation method in \cite{sui2023novel} is a conceptual advance, its multi-second convergence time remains significantly slower than human performance and is insufficient for real-time applications.

A fundamental limitation underlying above robotic grasping strategies is the decoupling of friction estimation from grasp control. In contrast, human grasping demonstrates a tightly synchronized process where haptic feedback and motor control are seamlessly integrated to achieve reliable manipulation \cite{johansson2009coding}. To endow robots with similar adaptive capabilities, two primary challenges must be addressed: (1) achieving rapid and robust online friction estimation and (2) developing a control framework that seamlessly integrates this estimation in real time. Therefore, we frist proposed a synchronized strategy to achieve the continuous integration of sensing and control that characterizes human sensorimotor coordination.

In contrast to treating friction as a fixed, deterministic parameter, we model it as a stochastic variable. Therefore, we define the system state as $\mu_t$. Its state evolution is modeled as a random walk process:

\begin{align}
\label{eq-system model}
    \mu_{t} &= \mu_{t-1} + \epsilon_{t}, \quad \epsilon_{t} \sim \mathcal{N}(0, \sigma_{p}^{2})\\
    \mu_{t} &= \text{min}(\text{max}(\mu_{t} \mu_{min}), \mu_{max})
\end{align}
where $\sigma_p^2$ represents process noise variance. Here, $\mu_{min}$ and $\mu_{max}$ are the physical constrains boundary. The state of interest is referred to as the posterior state, $p(\mu_{t} | z_{1:t})$, where $z_{1:t} = {z_1, z_2, \ldots, z_t }$ represents the sequence of all observations from the start of the grasp until time $t$. Then the particle filter is used to estimate the friction coefficient $\mu_{t}$ in the real time.

\begin{equation}
    p(\mu_{0:t} \mid z_{1:t}) \approx \sum_{i=1}^{M}w_{t}^{i} \delta (\mu_{0:t} - \mu_{0:t}^{i})
\end{equation}
where $\{w_{t}^{i}, \mu_{0:t}^{i}\}_{i=1}^{M}$ is the set of $M$ weighted samples. The weight $w_{t}^{i} \in (0, 1]$ signifies the relative importance of the sample $\mu_{0:t}^{i}$, with $\sum_{i=1}^{M}w_{t}^{i}=1$. Samples with higher weights correspond to more probable estimates of the true state sequence. Here, $\delta(\cdot)$ denotes the Dirac delta function, which is zero everywhere except at the origin and integrates to one.

The process of grasping, lifting, and replacing an object can be decomposed into seven distinct phases: reach, load, lift, hold, replace, unload, and release \cite{johansson2009coding}. During the reach phase, the objective is to establish contact between the object and the tactile sensor-equipped gripper. The gripper closes gradually until a contact is detected, which is determined by a threshold on the resultant normal force: $F_{n} > F_{n}^{\text{init}}$, where $F_{n}^{\text{init}}$ is the predicted initial force got from \model. 

Subsequently, during the load phase, the friction coefficient estimation enables the online update of the contact coefficient $cf_{t}$ via (\ref{eq-contact coefficient}), utilizing the measured forces. A proportional control law adjusts the gripper position $x_g$ to drive $cf_t$ towards a predefined target $cf_{\text{target}}$, which defines the desired safety margin against slip. The gripper command is given as following, and $K_p$ is a proportional gain.
\begin{equation}
\Delta x_g(t) = K_p \cdot (cf_{\text{target}} - cf_t)
\label{eq-control_law}
\end{equation}

It is noteworthy that the friction estimation module and the adaptive grasp controller form a closed-loop system: the controller utilizes the current best estimate to adjust the gripper position, while new tactile feedback resulting from this adjustment is continuously incorporated to refine the estimation. This reciprocal interaction establishes a highly responsive and disturbance-resistant sensorimotor cycle.

\subsection{Infrastructure as Enabler: Toward a Scalable and Expansive 
Roadmap}\label{sec:infra}

This section details our roadmap for operationalizing our theory, which is built upon two pillars of adjusted applications: a parallelization strategy (Context Parallelism) to scale context length and a custom training framework to conquer the challenges of training a 72B-scale multi-modal model with RL.

\paragraph{Context Parallelism: Scaling Multi-modal Context and Efficiency.} While DAPO optimizes the learning from each data point, the increasing complexity of multi-modal, long-horizon tasks necessitates processing ever-larger contexts, including long dialogue histories, sequential visual observations, and action histories. To overcome the memory bottlenecks of traditional data parallelism, we adopt and optimize Context Parallelism.

\textbf{Core Mechanism}: Context Parallelism addresses this challenge by distributing the input context (tokens, image patches) across multiple GPUs for a single forward pass.
\textbf{Split Context}: The input sequence is partitioned into smaller chunks, with each chunk assigned to a different GPU.
\textbf{Parallel Attention}: Each GPU computes the attention mechanism for its local chunk while communicating intermediate Key-Value (KV) states with other GPUs to maintain a global context. This allows the model's context window to scale linearly with the number of available devices, enabling the processing of vast temporal and spatial information crucial for embodied agents.

\paragraph{The Apex Challenge: Enabling RLHF on a 72B Multi-modal Model}. The ultimate ambition of our work is to build a 72-billion parameter multi-modal hybrid model. A significant gap in the current open-source ecosystem is the lack of robust frameworks capable of handling RL from Human Feedback (RLHF) for multi-modal models at such a scale.

\textbf{Our Solution}: To solve this, we adapt and significantly extend the VERL framework. Our primary contribution is the development of a novel training pipeline that enables a 72B-scale model to be trained with our preference learning algorithms on complex, mixed-modality data.
\textbf{Mixed-Batch Multi-modal Training}: The cornerstone of our innovation is a sophisticated batching and data loading mechanism that allows a single training batch to concurrently contain text-only, image-text, and video-text data. This heterogeneous batching capability is critical for efficient multi-modal learning.
\textbf{Scaling GRPO to 72B}: By leveraging our modified VERL framework and Context Parallelism, we successfully implement the GRPO algorithm at the 72B scale on these mixed-modality batches. This allows the model to learn from rich, ranked preference signals that span across different data types simultaneously, a capability that, to our knowledge, has not been demonstrated in open-source models of this magnitude.
\textbf{Conquering Engineering Hurdles}: This achievement required systematically solving a cascade of engineering challenges, including:
    Extreme Parallelism: Integrating a hybrid of data, pipeline, and tensor parallelism to manage the model's immense size.
    Multi-modal Fusion at Scale: Designing robust cross-attention mechanisms that effectively fuse information from disparate modalities without succumbing to training instability.
    Inference Optimization: Developing quantization and deployment strategies to make such a large model usable in practical applications.

By systematically addressing these infrastructural, algorithmic, and architectural challenges, our roadmap unlocks the full potential of large multi-modal models, providing a concrete and scalable path towards truly intelligent and capable embodied AI.
Our framework is built upon the unification of SFT and RL through the lens of preference learning, implemented via a multi-stage and iterative training pipeline.

\subsection{Reward Design for Embodied Tasks}

To accommodate the diverse modalities and objectives of embodied tasks, we design \textbf{task-specific reward functions}, all unified under a common interface for the RL framework.

\paragraph{(1) Format Reward.} 
The format reward ensures compliance with the required output pattern, enforcing consistent reasoning syntax without constraining intermediate ``thinking'' steps:
\begin{equation}
r_{\text{format}} =
\begin{cases}
1, & \text{if the output contains a valid } \\
0, & \text{otherwise.}
\end{cases}
\end{equation}
This soft structural constraint stabilizes learning and ensures that generated outputs are parsable for downstream evaluation.

\paragraph{(2) Choice Reward.}
For standard classification or multiple-choice questions, correctness is directly evaluated as:
\begin{equation}
r_{\text{choice}} = \mathbb{1}(\text{pred} = \text{label})
\end{equation}
which encourages precise logical discrimination and serves as a baseline for assessing embodied question-answering consistency.

\paragraph{(3) Numeric Reward (Counting / Distance).}
For tasks yielding quantitative outcomes—such as counting visible objects or estimating spatial distances—we employ a continuous reward reflecting numerical accuracy:
\begin{equation}
r_{\text{numeric}} = \exp(-\alpha |\hat{y} - y|)
\end{equation}
where $\hat{y}$ is the predicted numeric value, $y$ is the ground truth, and $\alpha$ controls sensitivity to error. 
This design rewards approximate correctness and provides smoother gradients for regression-like embodied reasoning.

\paragraph{(4) Affordance Reward.}
For affordance-centric tasks (e.g., determining which object can be grasped, pushed, or opened), we define:
\begin{equation}
r_{\text{affordance}} = \text{sim}(\text{pred}, \text{ground truth})
\end{equation}
where $\text{sim}(\cdot)$ can be instantiated as semantic matching or embedding-based alignment between the predicted and target affordance categories. 
This reward encourages the model to ground perception in actionable affordance semantics.
\subsection{Benhmark Analysis}
\begin{table*}[h]
\footnotesize
\centering
\caption{Task examples for each dimension in our benchmark.}
\begin{tabular}{l|l|l}
\toprule
\textbf{Primary dimensions}& \textbf{Secondary dimensions}              & \textbf{Example}      \\
\midrule
{Physical \& Causal }
&Physical feasibility            & Can the tissue box fit behind the chair? — Yes. \\
Reasoning&Stability and support            & Will the stacked cups remain stable if the top one is pushed slightly? — No. \\
&Intuitive physics and collisions           & If the rolling ball hits the block, will it fall over? — Yes. \\
\midrule
{Perception \& Object}
&Click or coordinate prediction             & Please point out the orange box. — (x, y) = (0.43, 0.72). \\
Grounding&Bounding box detection            & Identify the bounding box of the cup on the table. — (x1, y1, x2, y2). \\
&Grasp point           & Where should the robot grasp the handle of the kettle? — At the handle center. \\
\midrule
{Quantitative \& Numerical}
&Counting             & How many apples are on the table? — Three. \\
 Reasoning&Continuous estimation            & What is the distance between the box and the wall? — About 0.5 meters. \\
&Relative magnitude            & Which container holds more water? — The right one. \\
\midrule
{Spatial \& Geometric}
&Spatial relations             & Can the speaker fit left of the cup? — Yes. \\
 Reasoning&Egocentric vs allocentric frames            & From the robot’s view, which object is on its right? — The red box. \\
&Topological relations            & Which object is inside the basket? — The towel. \\
\midrule
{Temporal \& Sequential}
&Temporal order             & What happened before the man picked up the ball? — He bent down. \\
Reasoning&State transition and action effects            & After the cup is tilted, what happens to the water? — It spills. \\
&Motion and sequence reasoning            & What is the person doing after opening the door? — Walking inside. \\
\midrule
{Affordance \& Function}
&Graspability             & Which part of the screwdriver should be grasped? — The handle. \\
 Reasoning&Openability or pourability            & Can the bottle be opened by twisting the cap? — Yes. \\
&Tool-use reasoning            & What tool can be used to cut the rope? — Scissors. \\
&Manipulation feasibility           & Is it possible to rotate the knob with one hand? — Yes. \\
\midrule
{Multi-Object \& Scene}
&Multi-object relations             & Which object is partially hidden behind the monitor? — The keyboard. \\
 Consistency&Scene-state consistency           & Is the cup in the same place across both frames? — Yes. \\
&Spatial–language consistency           & Does the description match the image scene? — No. \\
\midrule
{Scene \& Action}
&Scene types            & What type of environment is this image captured in? — Kitchen. \\
 Understanding&Action recognition            & What is the person doing in the video? — Pouring water. \\
&State recognition            & Is the fridge door open or closed? — Open. \\
&Pose, gesture, direction            & What direction is the person pointing to? — Left. \\
\midrule
{Decision \& Task }
&Next-action prediction             & What should the robot do after picking up the cup? — Place it on the tray. \\
Planning&Sequential plan            & List the steps to make a cup of tea. — Boil water → Add tea → Pour → Serve. \\
&Sub-goal decomposition           & What are table assembly sub-goals? — Align legs → Screw bolts → Tighten. \\
\bottomrule
\end{tabular}

\label{tab:benchmark}
\end{table*}

\begin{table*}[h]
\small
\footnotesize
\centering
\caption{Coverage of the nine primary embodied reasoning dimensions across mainstream benchmarks and our proposed dataset.}
\setlength{\tabcolsep}{5pt} 
\renewcommand{\arraystretch}{1.3} 
\begin{tabular}{l|cccccccccc} 
\toprule
\textbf{Primary dimensions} &
\textbf{COSMOS} &
\textbf{ERQA} &
\textbf{Phyx} &
\textbf{Where2Place} &
\textbf{VSI-Bench} &
\textbf{BLINK} &
\textbf{RoboSpatial} &
\textbf{EmbSpatial} &
\\
\midrule
Physical \& Causal Reasoning & \checkmark & \checkmark & \checkmark &  & \checkmark & \checkmark & \checkmark & \checkmark \\
Perception \& Object Grounding &  & \checkmark &  & \checkmark &  & \checkmark & \checkmark &  \\
Quantitative \& Numerical Reasoning &  & \checkmark & \checkmark &  & \checkmark & \checkmark &  & \checkmark \\
Spatial \& Geometric Reasoning & \checkmark & \checkmark & \checkmark & \checkmark & \checkmark & \checkmark & \checkmark & \checkmark \\
Temporal \& Sequential Reasoning & \checkmark & \checkmark & \checkmark &  & \checkmark & \checkmark &  &   \\
Affordance \& Function Reasoning & \checkmark & \checkmark & \checkmark & \checkmark & \checkmark & \checkmark &  &   \\
Multi-Object \& Scene Consistency & \checkmark & \checkmark & \checkmark &  & \checkmark & \checkmark &  & \checkmark  \\
Scene \& Action Understanding & \checkmark & \checkmark & \checkmark &  & \checkmark & \checkmark & \checkmark & \checkmark  \\
Decision \& Task Planning & \checkmark & \checkmark &  &  & \checkmark &  &  &  \\
\bottomrule
\end{tabular}
\label{tab:coverage}
\end{table*}

\end{document}